\documentclass{article}

\usepackage{arxiv}
\usepackage{longtable}
\usepackage[utf8]{inputenc} 
\usepackage[T1]{fontenc}    
\usepackage{hyperref}       
\usepackage{url}            
\usepackage{booktabs}       
\usepackage{amsfonts}       
\usepackage{nicefrac}       
\usepackage{microtype}      
\usepackage{lipsum}		
\usepackage{graphicx}
\usepackage{natbib}
\usepackage{doi}
\usepackage{multirow}
\usepackage{adjustbox}
\usepackage{booktabs}
\usepackage{makecell}

\newcommand{\benchmarkname}{LabOSBench}

\title{LabOSBench: Benchmarking Computer Use Agents towards Scientific Instrument Control}

\author{Anqi Zou$^{1,2}$, Han Deng$^{1,3}$, Chengyu Zhang$^{1}$, Junquan Hu$^{1,2}$, Yu Wang$^{1,2}$, Yuxiang Xing$^{1,2}$, \\
\textbf{Aokai Zhang$^{1,2}$, Hanling Zhang$^{1,3}$, Zhaoyang Liu$^{\dagger,4}$, Ben Fei$^{\dagger,3}$, Zhihui Wang$^{\dagger,2}$, Wanli Ouyang$^{1,3}$}
\\
$^1$ Shenzhen Loop Area Institute\\
$^2$ Dalian University of Technology\\
$^3$ The Chinese University of Hong Kong\\
$^4$ The Hong Kong University of Science and Technology\\
\texttt{feynben@gmail.com, zyliumy@gmail.com, zhwang@dlut.edu.cn} \\
}

\date{}

\renewcommand{\shorttitle}{LabOSWorld}

\hypersetup{
pdftitle={A template for the arxiv style},
pdfsubject={q-bio.NC, q-bio.QM},
pdfauthor={David S.~Hippocampus, Elias D.~Striatum},
pdfkeywords={First keyword, Second keyword, More},
}

\begin{document}
\maketitle

\begin{abstract}
Current computer-use benchmarks primarily focus on software operation tasks in virtualized systems, whereas scientific instrumentation scenarios require coordinated control over complex interfaces, and feedback-driven parameter adjustment.
However, directly evaluating agents on physical high-precision instruments is impractical due to high cost, safety risks, limited accessibility, and difficulty in ensuring reproducible evaluation.
This motivates the need for a simulated yet realistic testbed that preserves the operational challenges of scientific instruments while enabling scalable and safe benchmarking.
To this end, we introduce \textbf{\benchmarkname{}}, a challenging benchmark for multimodal GUI agents built on a suite of web-based scientific-instrument simulators.
Operating directly via a browser, \benchmarkname{} avoids resource-heavy OS virtualization while supporting flexible task configuration and execution-based evaluation.
Specifically, \benchmarkname{} constructs 96 subtasks across eight instrument simulators, covering workflows from sample loading, alignment, parameter tuning, and data acquisition to result inspection.
We evaluate general-purpose vision-language models, specialized GUI agent models, and advanced agentic frameworks at both subtask and end-to-end levels.
Our experiments reveal that while existing agents can complete many structured GUI subtasks, they still struggle with feedback-driven operations and long-horizon workflow execution.
Overall, \benchmarkname{} provides a reproducible, low-cost testbed for advancing computer-using agents toward scientific-instrument control.
\end{abstract}

\section{Introduction}

Humans perform many computational tasks through graphical user interfaces (GUIs) and command-line interfaces (CLIs), including web browsing, file management, data analysis, and software development. Recent advances in large vision-language models (VLMs) have enabled autonomous digital agents that follow natural-language instructions and interact with GUIs through reasoning-and-acting loops~\citep{yao2022react, qin2025ui}, offering a promising path toward simplifying human--computer interaction.

Despite this progress, developing multimodal agents for domain-specific professional settings remains challenging. Existing benchmarks such as \textit{OSWorld} provide full operating-system environments~\citep{xie2024osworld,bonatti2024windows,rawles2025androidworld}, but rely on heavy virtualization frameworks such as VMware or Docker, making them costly to deploy and difficult to scale. Web-based benchmarks such as \textit{WebArena} and \textit{Mind2Web} focus on general web navigation~\citep{zhou2024webarena,deng2023mind2web}, and thus do not capture scientific-instrument interfaces with intricate layouts, long-horizon procedures, and continuous parameter tuning. Moreover, many prior benchmarks rely on static demonstrations or offline evaluation, limiting their ability to assess interactive learning, exploration, and alternative valid action sequences.

Scientific instrument interfaces pose unique challenges for multimodal agents. Benchmarking this setting is difficult because it requires realistic procedural workflows, dense professional interfaces, and feedback-driven parameter tuning, while avoiding the cost, safety risks, and limited accessibility of physical instruments. Although recent work has explored scientific agents, autonomous laboratories, and intelligent instrumentation systems~\citep{tom2024self,szymanski2023autonomous,boiko2023autonomous,wang2022scienceworld,jansen2024discoveryworld,deng2026owl}, existing benchmarks provide limited support for evaluating GUI-based instrument control. Prior work mainly focuses on scientific reasoning, experiment planning, autonomous discovery, or instrument-specific intelligence~\citep{tom2024self,szymanski2023autonomous,boiko2023autonomous,wang2022scienceworld,jansen2024discoveryworld}, whereas our work evaluates existing computer-use agents on scientific-instrument GUIs when direct device APIs are unavailable.

To address this gap, we introduce \textbf{\benchmarkname{}}, a lightweight, web-based, executable benchmark for evaluating multimodal agents on scientific instrument simulators. \benchmarkname{} runs entirely in a standard browser environment, eliminating the need for OS-level virtualization while preserving realistic GUI interactions. It supports native mouse and keyboard control, configurable initial states, execution-based evaluation.

\benchmarkname{} includes 96 subtasks across eight instrument simulators, covering workflows such as X-ray diffraction scanning, focused ion beam milling, light/fluorescence microscopy imaging, and scanning probe microscopy operation. We evaluate state-of-the-art LLM/VLM-based agents, specialized GUI models~\citep{qin2025ui}, and agentic frameworks~\citep{han2026vlaa}. Results show that strong multimodal models can complete many structured subtasks, but still struggle with feedback-driven adjustment, scientific-state interpretation, and long-horizon execution. Our analysis further reveals failures in visual grounding, action localization, recovery strategy, and instrument-specific GUI understanding.

Our contributions are threefold: (i) we introduce \benchmarkname{}, a lightweight executable benchmark for scientific-instrument GUI control; (ii) we design 96 subtasks across eight instrument simulators, covering procedural operation, visual feedback, and scientific-state adjustment; and (iii) we systematically evaluate general-purpose multimodal models, specialized GUI models, and agentic frameworks, revealing substantial gaps in feedback-driven and end-to-end scientific workflows.

\begin{figure*}[t]
  \centering
  \includegraphics[width=\textwidth]{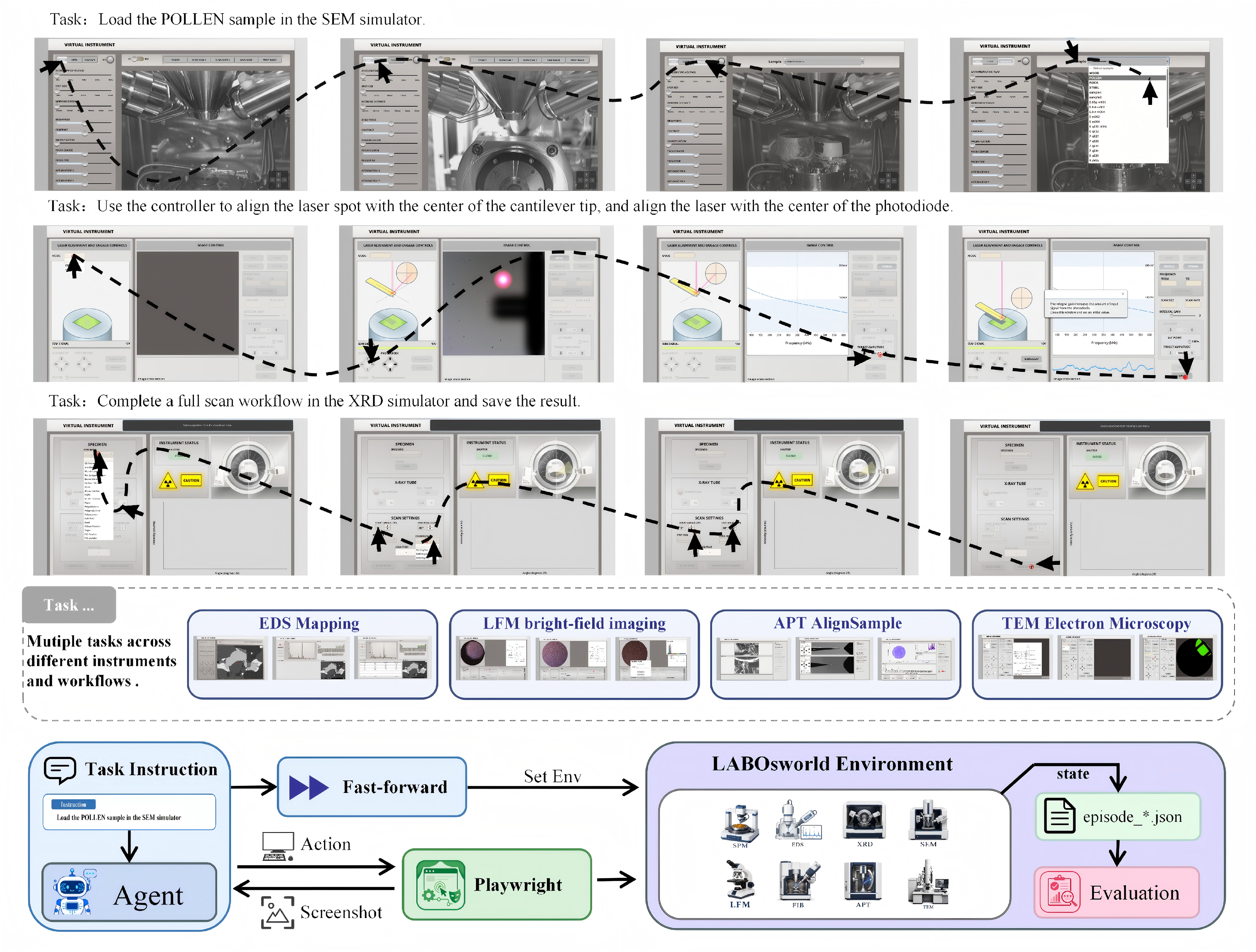}
  \caption{Overview of \benchmarkname{}. The benchmark evaluates computer use agents on eight types of scientific instrument web interfaces, where agents receive natural-language instructions and screenshots, execute GUI actions, and are scored using episode-level and subtask-level metrics across 96 subtasks.}
  \label{fig:overview}
\end{figure*}

\section{Related Work}

\paragraph{Computer-Using Agents (CUAs).}
CUAs have evolved from DOM-based web agents to vision-based computer-use agents that operate directly on screenshots~\citep{deng2023mind2web,zhou2024webarena,he2024webvoyager,zheng2024gpt,xie2024osworld,davydova2025osuniverse,rawles2025androidworld,qin2025ui,xu2025mobilerlonlineagenticreinforcement,liu2025scalecua,wu2025oracle,li2026themis}. Screen parsing methods~\citep{lu2024omniparser,li2025screenspot} improve agents' ability to identify actionable elements in complex visual observations. However, current agents still struggle with long-horizon planning, precise localization, and state-dependent interaction dynamics.

\paragraph{Agents for Science.}
Scientific-agent systems have studied reasoning, experiment planning, chemistry tool use, machine-learning experimentation, and autonomous discovery~\citep{tom2024self,szymanski2023autonomous,boiko2023autonomous,wang2022scienceworld,jansen2024discoveryworld,bran2023chemcrow,huang2023mlagentbench,lu2024ai}. MyScope\footnote{\url{https://myscope.training/}} provides browser-based microscope simulators for human education. In contrast, \benchmarkname{} adds executable evaluation for computer-use agents, including natural-language task specifications, subtask-level success checkers, episode logging, browser execution, and instrument-specific metrics.

\paragraph{Computer-Using Benchmarks.}
Existing benchmarks evaluate agents in web, desktop, mobile, and scientific workflow environments~\citep{deng2023mind2web,zhou2024webarena,koh2024visualwebarena,xie2024osworld,xu2024androidlabtrainingsystematicbenchmarking,davydova2025osuniverse,rawles2025androidworld,yang2026macosworld,sun2025scienceboard}. Web-based benchmarks are lightweight but focus mainly on general navigation, while OS-level benchmarks improve realism at the cost of heavy virtualization. More recently, \textit{ScienceBoard}~\citep{sun2025scienceboard} evaluates multimodal autonomous agents in realistic scientific workflows with professional software. However, existing benchmarks still provide limited coverage of scientific-instrument GUI operation, where agents must manipulate dense control panels, track instrument states, perform continuous parameter adjustment, and respond to state-dependent visual or numerical feedback. In contrast, \benchmarkname{} focuses on executable, browser-based scientific instrument simulators to evaluate GUI-based instrument control when device APIs are unavailable.

\section{The LabOSWorld Benchmark}
\label{sec:benchmark}

We introduce \benchmarkname{}, a benchmark suite for evaluating computer-use agents on scientific-instrument interfaces. Unlike general web or desktop tasks, scientific-instrument control requires procedural dependencies, domain-specific controls, state transitions, and ordered acquisition or export operations. \benchmarkname{} complements general computer-use benchmarks~\citep{xie2024osworld,zhou2024webarena,deng2023mind2web,rawles2025androidworld} with domain-specific web interfaces and laboratory-style workflows. 

\begin{figure*}[t]
  \centering
  \includegraphics[width=0.96\textwidth]{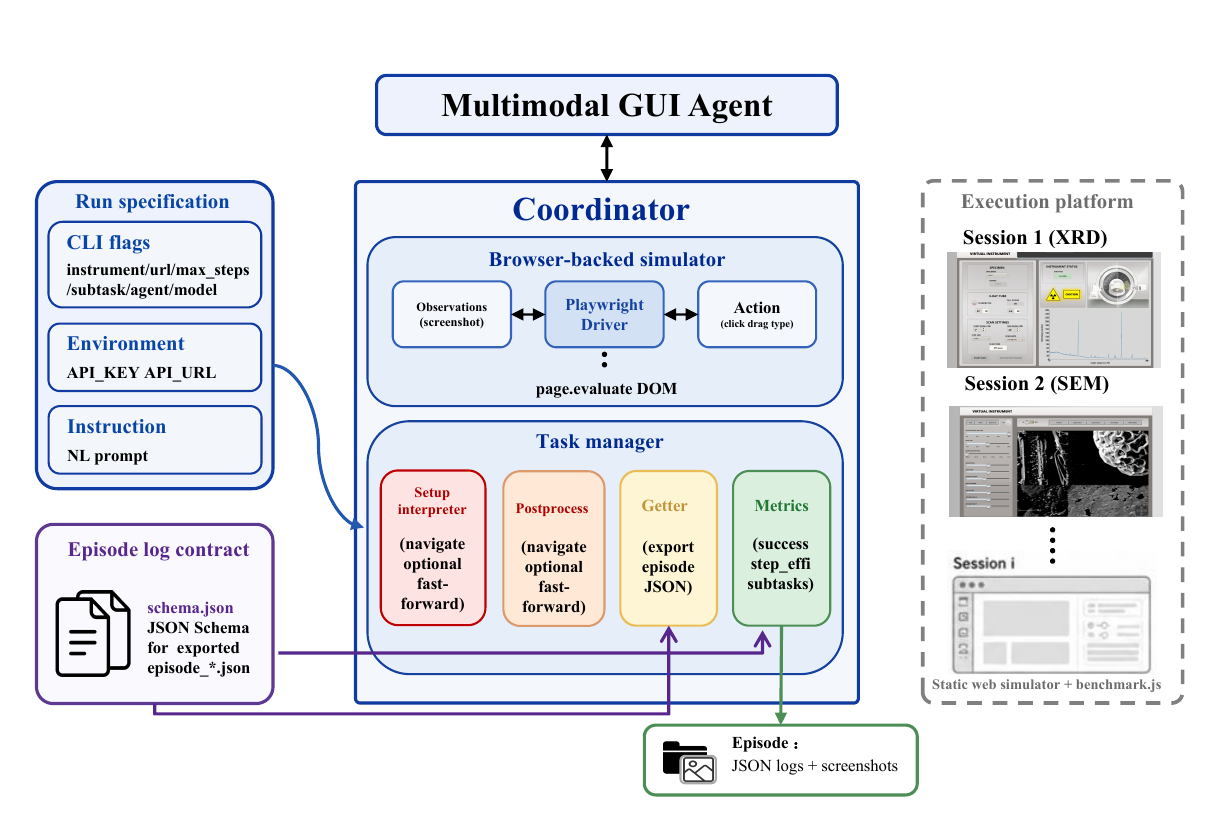}
  \caption{Evaluation infrastructure of \benchmarkname{}. Each run is specified by command-line arguments, environment variables, and a natural-language instruction. A browser-backed coordinator executes agent actions in scientific instrument web simulators, exports episode logs through in-page benchmark hooks, and aggregates metrics according to instrument-specific JSON Schemas.}
  \label{fig:infra}
\end{figure*}

Each run is launched from a lightweight specification consisting of command-line arguments, environment variables, and a natural-language instruction. As shown in Fig.~\ref{fig:infra}, a browser-backed coordinator executes agent actions, exports episode logs through in-page benchmark hooks, and aggregates instrument-specific metrics.

\subsection{Scientific Instrument Tasks}
\label{sec:tasks}

\benchmarkname{} covers eight simulated scientific instruments that span three foundation experimental paradigms: 
(i) advanced microscopy imaging, including Scanning Electron Microscopy (\textbf{SEM}), Transmission Electron Microscopy (\textbf{TEM}), and Light/Fluorescence Microscopy (\textbf{LFM});
(ii) Diffraction and Spectroscopic Analysis, including X-Ray Diffraction (\textbf{XRD}), Energy-Dispersive Spectroscopy (\textbf{EDS}) and Atom Probe Tomography (APT);
(iii) precision micro-nanofacrication and manipulation, including Focused Ion Beam (\textbf{FIB}) and Scanning Probe Microscopy (\textbf{SPM}). 
Each instrument is implemented as a high-fidelity, interactive web environment that meticulously preserves domain-specific control modalities, such as multi-dimensional sliders, discrete toggles, real-time dynamic plots, and multi-state status indicators.

Each task within our benchmark is operationalized via a natural-language instruction describing the target scientific operation.
Rather than navigating standardized, linear web forms, agents in LabOSWorld must master highly heterogeneous, instrument-specific operational logic.
For example, XRD tasks require agents to select a specimen, configure scan parameters, run the scan, and save the diffraction result, while SEM tasks involve chamber control, sample selection, vacuum preparation, high-voltage activation, imaging adjustment, scanning, and image saving. 
Consequently, successfully solving these tasks demands an integration of fine-grained cross-modal visual-grounding, non-linear procedural reasoning under physical dependencies, and robust state tracking across extended interaction horizons.

\subsection{Workflow and Subtask Decomposition}
\label{sec:subtasks}

To facilitate structured evaluation, we structure the workflow of each instrument as a sequence of subtasks.
Each subtask corresponds to a meaningful operation stage detectable from simulator states or interactions with specific GUI controls. 
This decomposition follows actual scientific operating procedures rather than arbitrary or heuristic screen-region segmentations.

\begin{figure*}[t]
  \centering

  \begin{minipage}[c]{0.56\textwidth}
    \centering
    \includegraphics[
      width=0.84\linewidth
    ]{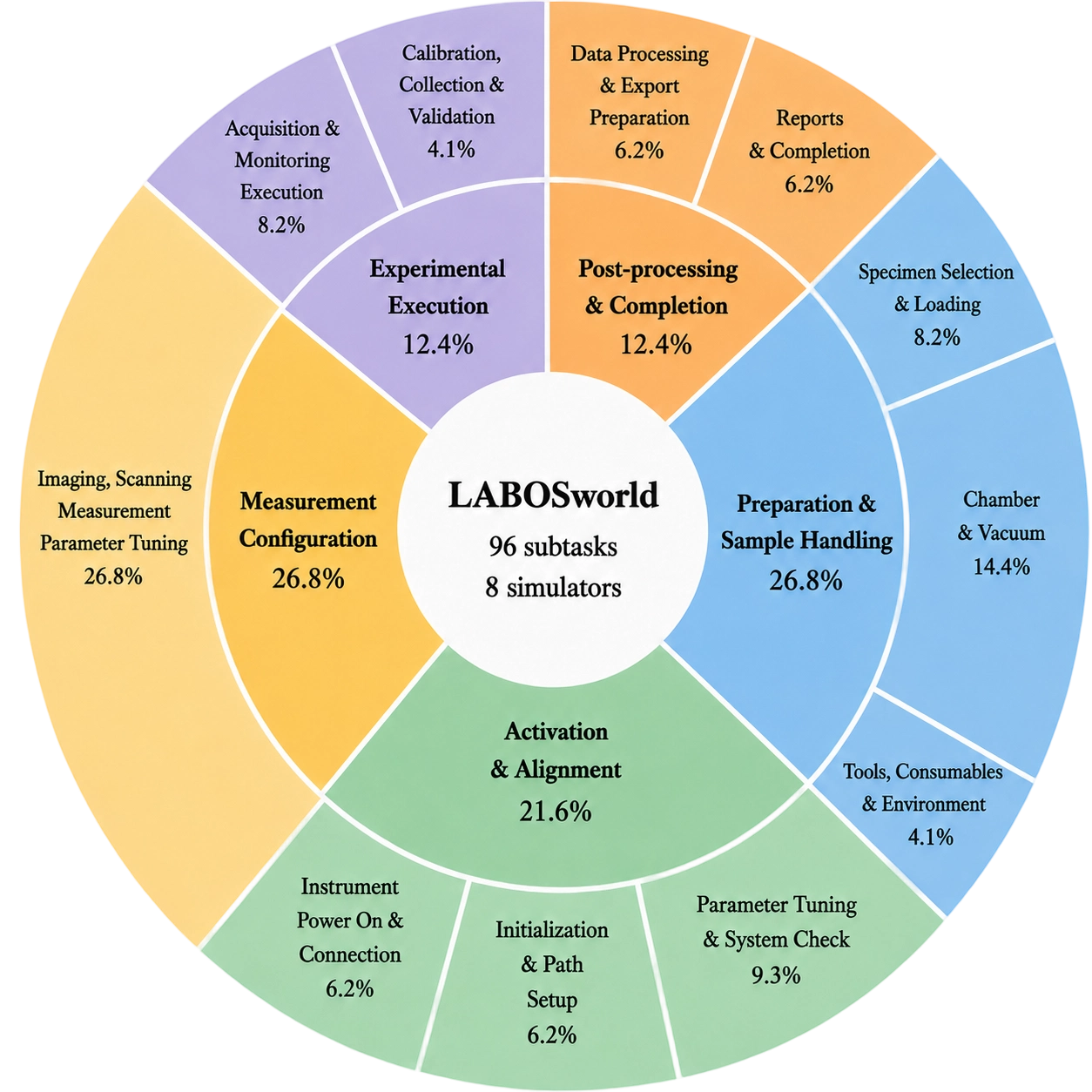}
  \end{minipage}
  \hspace{0.02\textwidth}
  \begin{minipage}[c]{0.36\textwidth}
    \centering
    \resizebox{0.90\linewidth}{!}{%
      \begin{tabular}{lcl}
        \toprule
        \textbf{Inst.} & \textbf{\#Sub.} & \textbf{Goal} \\
        \midrule
        XRD & 8  & Diffractogram \\
        SEM & 12 & Imaging \\
        TEM & 10 & Imaging \\
        FIB & 20 & Preparation \\
        SPM & 14 & Scanning \\
        LFM & 12 & Brightfield \\
        EDS & 8  & Spectrum \\
        APT & 12 & Reconstruction \\
        \midrule
        \textbf{Total} & \textbf{96} & -- \\
        \bottomrule
      \end{tabular}
    }
  \end{minipage}

  \caption{Overview of scientific-instrument workflows and GUI task taxonomy in \benchmarkname{}. The left panel summarizes the distribution of 96 subtasks across high-level and fine-grained GUI operation categories, while the right panel lists the eight instrument simulators, their subtask counts, and workflow goals.}
  \label{fig:task_taxonomy_and_workflows}
\end{figure*}

Across instruments, subtasks fall into recurring structural categories:
(i) \textit{Sample preparation} (e.g., specimen selection and mounting);
(ii) \textit{Environment conditioning} (e.g., vacuum evacuation);
(iii) \textit{Power or beam activation} (e.g., power-up or beam alignment);
(iv) \textit{Parameter configuration} (e.g., focus adjustment and astigmatism correction);
(v) \textit{Data acquisition} (e.g., sample scanning and signal capturing); 
This design enables interpretable evaluation beyond binary task success and supports targeted diagnosis of failures at specific workflow stages.

\subsection{Browser-Based Evaluation Infrastructure}

Unlike traditional environments that depend on heavy virtual machines, \benchmarkname{} runs natively in a browser. 
We use Playwright to open simulator pages, capture screenshots, and execute agent actions. At each step, the agent receives the instruction and current screenshot, then outputs an executable GUI action. 
The action space covers clicking, double-clicking, dragging, typing, selecting dropdown options, pressing keys, scrolling, and waiting for simulator state transitions. 
The browser driver parses and executes these actions, using DOM-level event dispatch when possible and mouse events otherwise for JavaScript-based controls.

Each simulator page is instrumented with an in-page benchmark script, such as \texttt{benchmark\_xrd.js} or \texttt{benchmark\_sem.js}. These scripts record subtask completion, control attempts, episode metadata, and step-level traces.
Finally, a Python-based coordinator exports the in-page episode object and step screenshots, following instrument-specific JSON Schemas with fields such as \texttt{success}, \texttt{summary}, \texttt{subtasks}, and \texttt{steps}.

\subsection{Full-Episode and Subtask-Level Evaluation}
\label{sec:modes}

We evaluate agents in two complementary modes. In the full-episode setting, the agent starts from the initial simulator state and must complete the entire workflow within a pre-defined step budget. This setting measures long-horizon task completion and captures compounding failures caused by early mistakes, missed state transitions, or incorrect parameter settings.

In the subtask-level setting, the simulator is initialized directly to the canonical state immediately preceding a target subtask, and the agent is evaluated solely on that individual subtask. 
This mode isolates local capabilities such as locating a domain-specific widget, selecting a menu item, dragging a parameter slider, or recognizing operation completion. 
For simulators that support programmatic state initialization, such as XRD, we implement fast-forward functions exposed on the browser window, e.g., \texttt{XRD\_fast\_forward\_to\_subtask($S_k$)}. 
These functions mark preceding subtasks as completed and position the UI at the beginning of the target subtask. 
Such fast-forwarding is utilized exclusively for diagnostic subtask evaluation and is disabled during full-episode evaluation.

Together, the two settings distinguish local GUI grounding failures from long-horizon workflow failures. 
An agent that succeeds on subtasks but fails in full episodes may suffer from limitations in memory, sequential ordering, or error recovery, whereas failure on individual subtasks indicates weak grounding of domain-specific scientific controls.

\subsection{Metrics}
\label{sec:metrics}

We report episode-level and subtask-level metrics. An episode is successful if all required subtasks are completed within the step budget. For each subtask, we record whether the corresponding operation is completed, revealing where an agent fails in full episodes and enabling localized success-rate analysis.

Let $\mathcal{S}_d$ denote the set of subtasks associated with instrument $d$, and let $R_{d,s}$ denote the number of repeated runs available for subtask $s$. In our main subtask-level evaluation, $R_{d,s}=2$. For each subtask $s \in \mathcal{S}_d$, we compute:
\[
\mathrm{SR}_{d,s} =
\frac{1}{R_{d,s}}\sum_{r=1}^{R_{d,s}}
\mathbf{1}\left[\mathrm{success}_{d,s,r}\right],
\]
where $\mathrm{success}_{d,s,r}$ indicates whether the agent successfully completes subtask $s$ of instrument $d$ in run $r$.

The instrument-level score is then computed by averaging over all subtasks:
\[
\mathrm{Score}_d =
\frac{1}{|\mathcal{S}_d|}
\sum_{s \in \mathcal{S}_d}
\mathrm{SR}_{d,s}.
\]
The reported results in Table~\ref{tab:main_score_by_instrument} correspond to $\mathrm{Score}_d$ for each instrument. We also log interaction attempts and grounding-oriented statistics for diagnostic analysis.
For image-producing tasks, we compute pixel-level quality metrics when a reference image or canonical view is available. In SEM, the saved or current micrograph is compared against a reference view after display-region normalization. We report peak signal-to-noise ratio (PSNR) to measure the visual fidelity of the resulting micrograph. This metric assesses whether parameter adjustments produce a scientifically usable visual state beyond merely triggering the correct GUI action. 

\section{Experiments}

In this section, we present the experimental settings and main results for representative GUI agent baselines on \benchmarkname{}, including general-purpose multimodal models, specialized GUI models, and agentic frameworks.

\subsection{Baseline Models and Agentic Frameworks}
\label{sec:baselines}

We evaluate three groups of baselines on \benchmarkname{}. 
The first group includes general-purpose multimodal models, including Qwen3VL-32B~\citep{bai2025qwen3}, EvoCUA-8B~\citep{xue2026evocua}, Claude Sonnet-4.5~\citep{claude45sonnet}, Kimi-K2.5~\citep{team2026kimi}, Seed-1.6~\citep{seed16}, GPT-5.5~\citep{gpt55systemcard}, and Claude Opus-4.5~\citep{claude45opus}. 
The second group consists of specialized GUI models, including UI-TARS-1.5-7B~\citep{qin2025ui} and GUI-Owl-7B~\citep{deng2026owl}. 
The third group contains agentic frameworks built on strong foundation models, including GTA1 w/ GPT-5.5, VLAA-GUI w/ Opus-4.5~\citep{han2026vlaa}, and Hippo Agent w/ Opus-4.5.

For each task, the agent receives a natural-language instruction and the current screen observation, and then outputs an executable GUI action, such as clicking, typing, scrolling, selecting a widget, or triggering an instrument-specific operation. 
We use a unified interaction protocol across all instruments and models. 
For subtask-level evaluation, each model is given at most 50 interaction steps per subtask. 
Unless otherwise specified, we run each model--subtask pair twice (\texttt{runs=2}) and report subtask-normalized success rates.

For each instrument $d$, we first compute the success rate of each subtask over repeated runs and then average across all subtasks, ensuring that each subtask contributes equally. 
The instrument-level scores reported in Table~\ref{tab:main_score_by_instrument} correspond to this subtask-normalized score.

\begin{table*}[t]
\centering
\scriptsize
\setlength{\tabcolsep}{2.3pt}
\caption{Main results of average subtask-level scores on the \benchmarkname{} benchmark across eight scientific instrument simulators. Scores are reported in the range $[0,1]$. The Avg. column reports the macro-average over the eight instrument simulators. Bold values indicate the best non-human method for each column.}
\label{tab:main_score_by_instrument}
{%
\begin{tabular}{llccccccccc}
\toprule
\textbf{Category} & \textbf{Model}
& \textbf{SEM}
& \textbf{SPM}
& \textbf{TEM}
& \textbf{XRD}
& \textbf{LFM}
& \textbf{FIB}
& \textbf{APT}
& \textbf{EDS}
& \textbf{Avg.} \\
\midrule
\multirow{7}{*}{\makecell[l]{General\\Models}}
& Qwen3VL-32B
& 0.833 & 0.643 & 0.500 & 0.750 & 0.500 & 0.275 & 0.667 & 0.750 & 0.615 \\
& EvoCUA-8B
& 0.833 & 0.571 & 0.500 & 0.750 & 0.500 & 0.350 & 0.708 & 0.562 & 0.597 \\
& Claude Sonnet-4.5
& 0.833 & 0.714 & 0.400 & 0.625 & 0.500 & 0.250 & 0.667 & \textbf{1.000} & 0.624 \\
& Kimi-K2.5
& 0.625 & 0.357 & 0.500 & \textbf{0.875} & 0.667 & 0.525 & 0.667 & 0.562 & 0.597 \\
& Seed-1.6
& 0.792 & 0.786 & 0.650 & \textbf{0.875} & 0.667 & 0.650 & 0.875 & 0.812 & 0.763 \\
& GPT-5.5
& 0.833 & 0.786 & 0.500 & 0.625 & 0.500 & 0.650 & \textbf{0.917} & \textbf{1.000} & 0.726 \\
& Claude Opus-4.5
& 0.833 & 0.643 & 0.400 & 0.750 & 0.500 & 0.250 & 0.583 & \textbf{1.000} & 0.620 \\
\midrule
\multirow{2}{*}{\makecell[l]{Specialized\\Models}}
& UI-TARS-1.5-7B
& 0.500 & 0.750 & 0.600 & 0.812 & 0.542 & 0.650 & 0.667 & 0.750 & 0.659 \\
& GUI-Owl-7B
& 0.833 & 0.714 & 0.500 & 0.750 & 0.333 & 0.325 & 0.667 & \textbf{1.000} & 0.640 \\
\midrule
\multirow{3}{*}{\makecell[l]{Agentic\\Frameworks}}
& GTA1 w/ GPT-5.5
& \textbf{0.875} & \textbf{0.821} & \textbf{0.850} & \textbf{0.875} & \textbf{0.708} & \textbf{0.675} & 0.833 & 0.875 & \textbf{0.814} \\
& VLAA-GUI w/ Opus-4.5
& 0.333 & 0.429 & 0.400 & 0.625 & 0.333 & 0.150 & 0.500 & 0.875 & 0.456 \\
& Hippo Agent w/ Opus-4.5
& 0.833 & 0.714 & 0.500 & 0.750 & 0.583 & 0.300 & 0.583 & \textbf{1.000} & 0.658 \\
\midrule
\multirow{1}{*}{Human}
& Human
& 0.917 & 0.857 & 0.900 & 1.000 & 0.917 & 0.800 & 1.000 & 1.000 & 0.924 \\
\bottomrule
\end{tabular}%
}
\end{table*}

\begin{figure}[t]
  \centering
  \includegraphics[width=0.6\linewidth]{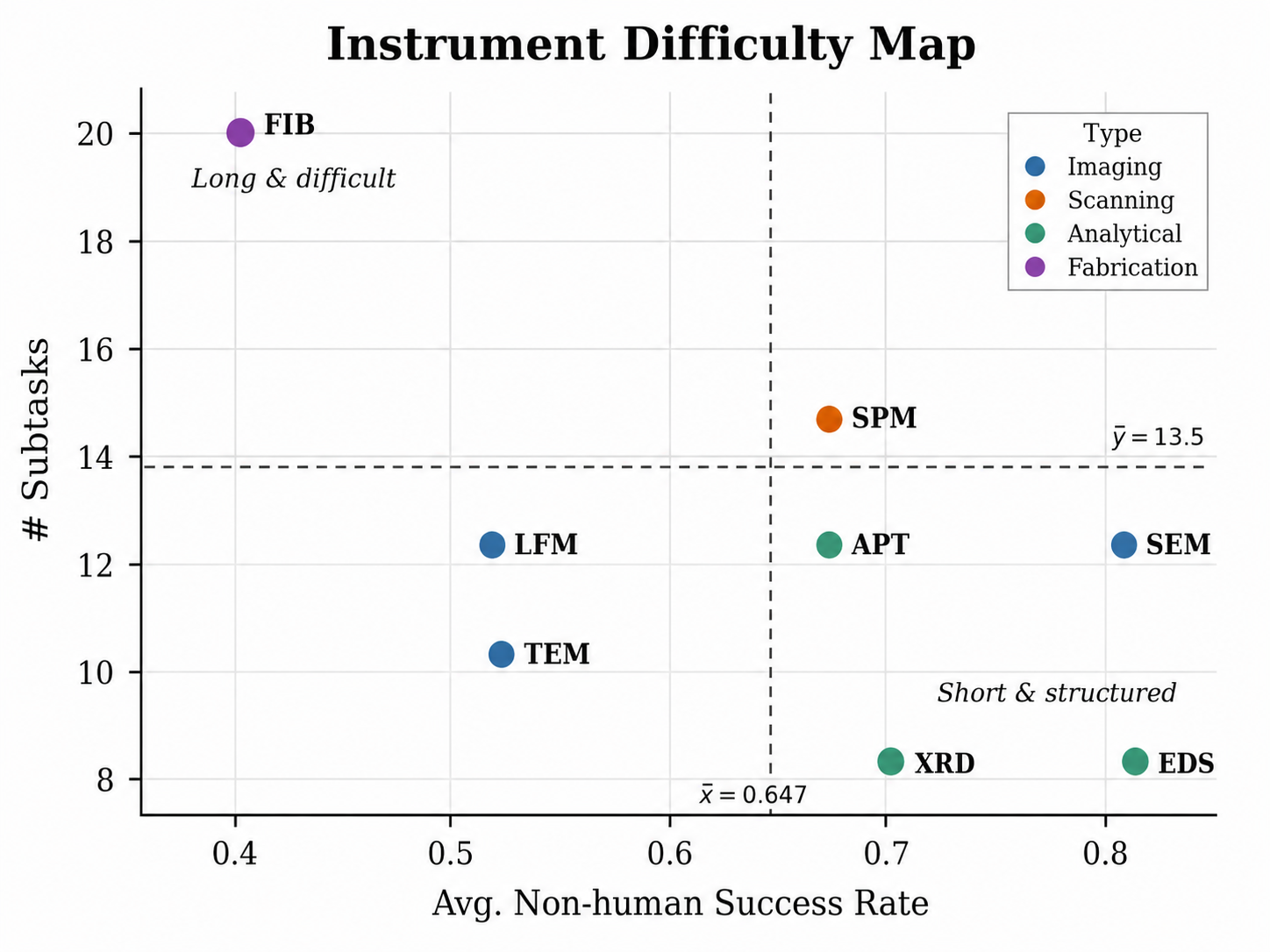}
  \vspace{-4pt}
  \caption{Instrument difficulty map of the eight scientific-instrument simulators in \benchmarkname{}. The x-axis shows the average non-human success rate, the y-axis shows the number of subtasks, and bubble size denotes the human--model gap. Dashed lines indicate instrument-level means.}
  \label{fig:instrument_difficulty_map}
\end{figure}

\paragraph{Benchmark-level difficulty.}
Before comparing individual models, we first analyze the structural difficulty of \benchmarkname{} at the instrument level. Figure~\ref{fig:instrument_difficulty_map} visualizes each simulator according to its average non-human model success rate, workflow length, and human--best model gap. This view shows that the difficulty of scientific-instrument GUI control is multi-dimensional. FIB is challenging because it combines the longest workflow with low average model performance, indicating strong long-horizon dependency and error-accumulation pressure. LFM has a shorter workflow than FIB but a larger human--model gap, suggesting that feedback-driven optical adjustment and visual-state interpretation remain difficult for current agents. In contrast, EDS and XRD are located in the short-workflow and high-performance region, reflecting their more structured operation sequences.

\subsection{Results and Analysis}
\label{sec:results_analysis}

\paragraph{Category-level analysis.}
Table~\ref{tab:category_success_rates} reveals distinct strengths and limitations across method families. General-purpose multimodal models benefit from strong visual perception and instruction following: Seed-1.6 performs well on experimental execution and post-processing, while GPT-5.5 is strong in measurement configuration and execution. This suggests that general VLMs can handle instruction understanding, visual feedback, and parameter selection, but still struggle with long-horizon tracking and fine-grained closed-loop adjustment.

Specialized GUI models show advantages in conventional interface grounding. UI-TARS-1.5-7B and GUI-Owl-7B are competitive on widget selection, panel operation, and structured GUI procedures. However, their gains are less consistent for tasks requiring scientific-state interpretation, such as alignment, focusing, or visually guided adjustment, indicating that GUI localization alone is insufficient for scientific-instrument control.

Agentic frameworks achieve the strongest overall results, but their benefits depend on reliable grounding and domain-valid recovery. GTA1 w/ GPT-5.5 obtains the best overall average and leads on Activation \& Alignment and Measurement Configuration, suggesting that planning, verification, and execution monitoring can improve complex workflows. In contrast, the weaker performance of VLAA-GUI w/ Opus-4.5 shows that simply adding an agent loop is not enough: wrong clicks, incorrect control adjustments, and invalid recovery actions can still derail scientific procedures.

\begin{table*}[t]
\centering
\scriptsize
\caption{Success rates across five scientific-instrument task categories. Scores are reported as percentages. Each category-level score is computed by averaging subtask-level success rates over all operation-level subtasks assigned to that category.}
\label{tab:category_success_rates}
{%
\begin{tabular}{lccccc}
\toprule
\textbf{Model}
& \textbf{\makecell{Preparation \&\\ Sample Handling}}
& \textbf{\makecell{Activation \&\\ Alignment}}
& \textbf{\makecell{Measurement\\ Configuration}}
& \textbf{\makecell{Experimental\\ Execution}}
& \textbf{\makecell{Post-processing \&\\ Completion}} \\
\midrule
Qwen3VL-32B
& 82.4 & 38.6 & 61.1 & 46.4 & 62.5 \\

Hippo Agent w/ Opus-4.5
& 88.2 & 43.2 & 66.7 & 42.9 & 65.6 \\

UI-TARS-1.5-7B
& 91.2 & 40.9 & 64.8 & 64.3 & 71.9 \\

GUI-Owl-7B
& 85.3 & 34.1 & 64.8 & 50.0 & 68.8 \\

EvoCUA-8B
& 85.3 & 40.9 & 64.8 & 50.0 & 43.8 \\

Kimi-K2.5
& 76.5 & 45.5 & 53.7 & 57.1 & 62.5 \\

Seed-1.6
& \textbf{97.1} & 59.1 & 66.7 & \textbf{82.1} & \textbf{81.2} \\

VLAA-GUI w/ Opus-4.5
& 82.4 & 18.2 & 29.6 & 28.6 & 56.2 \\

Claude Sonnet-4.5
& 76.5 & 45.5 & 63.0 & 35.7 & 68.8 \\

Claude Opus-4.5
& 82.4 & 40.9 & 63.0 & 35.7 & 62.5 \\

GTA1 w/ GPT-5.5
& 94.1 & \textbf{63.6} & \textbf{87.0} & 78.6 & 75.0 \\

GPT-5.5
& 82.4 & 50.0 & 81.5 & 71.4 & 75.0 \\
\bottomrule
\end{tabular}%
}
\end{table*}

\begin{figure}[t]
  \centering
  \includegraphics[width=0.6\columnwidth]{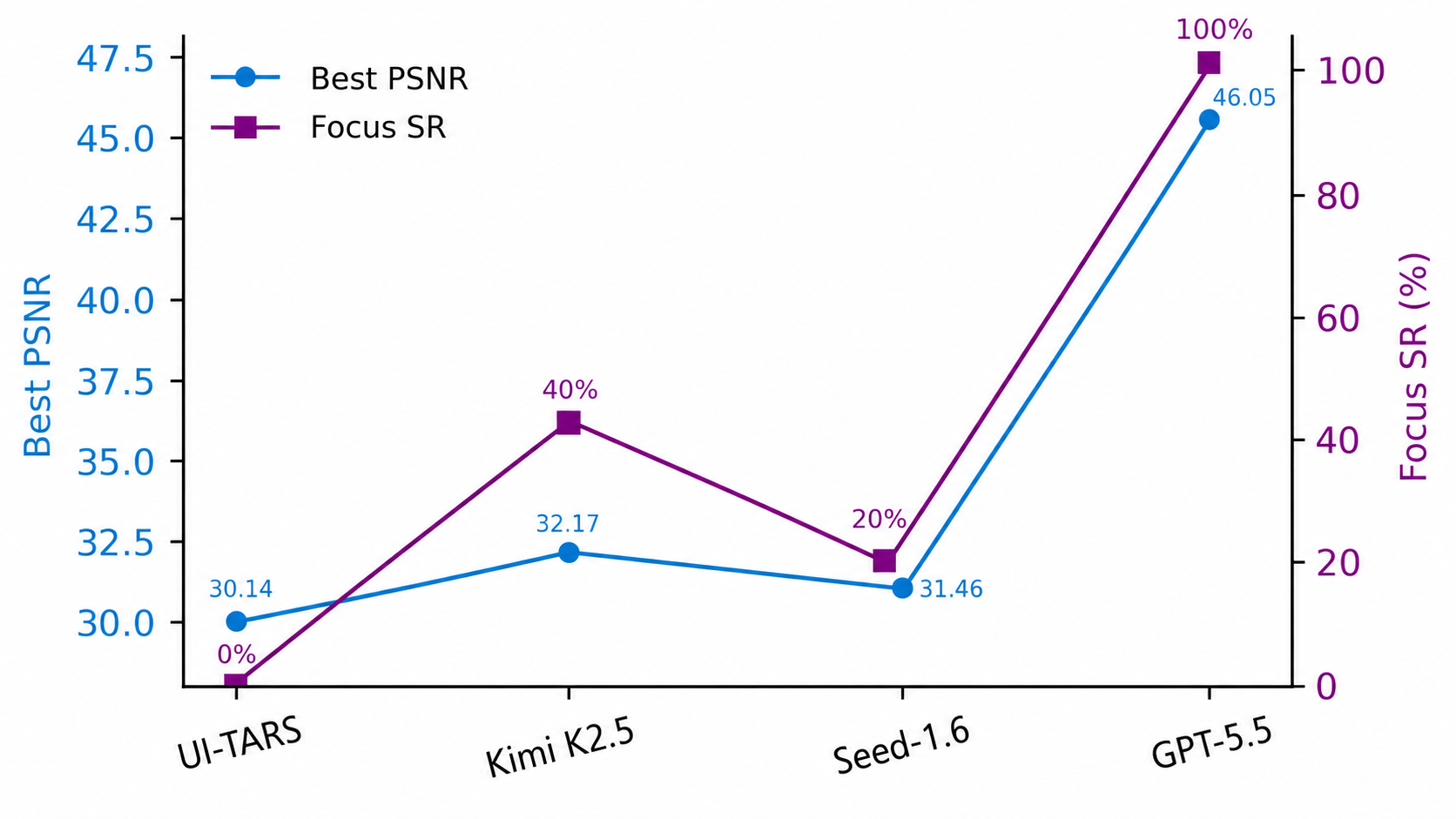}
  \vspace{-4pt}
  \caption{Scientific-state quality on the SEM focus-adjustment task. The two lines show best PSNR and focus success rate (SR) across models. GPT-5.5 achieves the highest values on both metrics.}
  \label{fig:sem_focus_lines}
\end{figure}

\paragraph{Scientific-state understanding.}
Table~\ref{tab:category_success_rates} suggests that scientific-instrument GUI control cannot be reduced to discrete widget grounding. Tasks involving alignment, focusing, and image adjustment require agents to interpret scientific visual states and perform feedback-driven control. We therefore analyze the SEM focus-adjustment task, where the agent must iteratively adjust focus and judge image clarity. As shown in Fig.~\ref{fig:sem_focus_lines}, GPT-5.5 achieves both the highest focus success rate and the highest best PSNR, showing that \benchmarkname{} can evaluate intermediate scientific-state quality rather than only whether a task is eventually completed.

\begin{figure}[t]
  \centering
  \includegraphics[width=0.6\columnwidth]{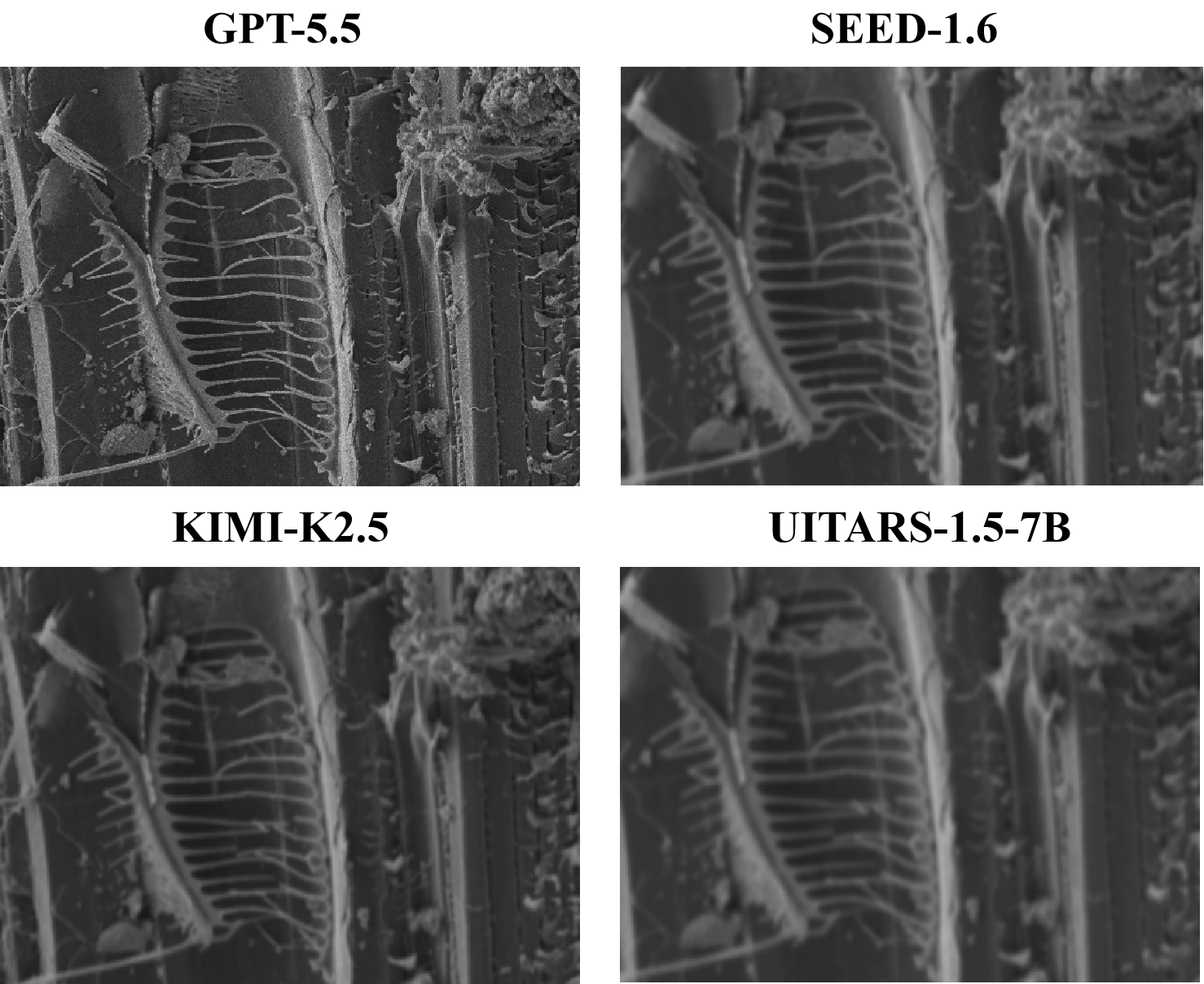}
  \vspace{-4pt}
  \caption{Qualitative comparison of final outputs on the SEM focus-adjustment task under the same task setting.}
  \label{fig:sem_focus_qualitative}
\end{figure}

\paragraph{End-to-end workflow difficulty.}
Although our main evaluation is subtask-level, we further perform an end-to-end pilot evaluation using GPT-5.5 with 10 runs per workflow. GPT-5.5 achieves non-zero success on EDS, APT, and XRD, with success rates of 70.0\%, 80.0\%, and 60.0\%, respectively, but fails on SEM, SPM, TEM, LFM, and FIB. The average end-to-end success rate is only 26.3\%, much lower than its subtask-level performance, indicating severe error accumulation in complete workflows. 

\paragraph{Instrument-level difficulty.}
Performance varies substantially across instruments. 
EDS and XRD are relatively easier due to structured operations, clear feedback, and short action sequences. 
In contrast, SEM, FIB, and LFM are more challenging because they involve feedback-driven focusing, image-quality adjustment, long procedures, fine-grained operations, and visual target localization. 
These differences show that \benchmarkname{} evaluates not only generic GUI navigation, but also visual grounding, scientific-domain understanding, long-horizon reasoning, and closed-loop control.

\paragraph{Summary.}
The results highlight four findings. 
First, subtask-level evaluation reveals limitations of current GUI agents, while end-to-end evaluation exposes severe error accumulation. 
Second, general-purpose multimodal models outperform specialized GUI models overall, but still struggle with feedback-intensive and long-horizon tasks. 
Third, agentic frameworks help only when their recovery strategies align with scientific-instrument operation semantics. 
Fourth, \benchmarkname{} challenges agents on GUI grounding, scientific-state understanding, sequential operation, and continuous visual adjustment.

\section{Conclusion}
\label{sec:conclusion}

We presented \benchmarkname{}, a lightweight executable benchmark for scientific-instrument GUI control built on web-based simulators. By avoiding full OS virtualization while preserving dense interfaces, procedural dependencies, continuous parameter tuning, and feedback-driven visual adjustment, \benchmarkname{} provides a practical testbed for evaluating multimodal GUI agents in scientific workflows. Our evaluation shows that current agents remain far from reliable scientific-instrument operation: strong multimodal models can solve many discrete subtasks, and agentic frameworks can improve robustness, but failures persist in scientific-state interpretation, precise spatial grounding, closed-loop adjustment, and long-horizon workflows. The SEM focus analysis and end-to-end study further demonstrate that successful agents must not only complete GUI actions, but also maintain valid intermediate scientific states and avoid error accumulation. Overall, \benchmarkname{} highlights scientific-instrument control as a challenging and underexplored setting that connects computer-use agent evaluation~\citep{xie2024osworld,zhou2024webarena,qin2025ui} with autonomous laboratory systems~\citep{tom2024self,szymanski2023autonomous,boiko2023autonomous}, and we hope it encourages future work on domain-aware GUI agents with accurate grounding, state-aware planning, and scientific feedback interpretation.

\section*{Limitations}

Although \benchmarkname{} provides a lightweight and executable testbed for scientific-instrument GUI control, it still has several limitations. First, it is built on web-based simulators rather than physical instruments, so it cannot fully capture hardware latency, calibration uncertainty, safety constraints, or real laboratory failure modes. Second, the current benchmark covers a selected set of instruments and workflows, including microscopy, spectroscopy, diffraction, and tomography, but does not yet include broader laboratory scenarios such as wet-lab protocols, robotic manipulation, chemical synthesis, or multi-instrument experimental planning. Finally, the evaluation mainly relies on screenshots and logged simulator states, while some scientific decisions may require richer observations, domain knowledge, or multimodal sensor signals beyond the current environment.

\benchmarkname{} is designed for evaluating agents in simulated scientific-instrument environments and does not provide direct control over real laboratory hardware. 
Nevertheless, scientific-instrument automation may involve safety-critical operations if deployed on physical devices. 
Future systems should include human supervision, permission control, operation logging, and safety interlocks before being connected to real instruments. 
The benchmark is intended to support reproducible research on GUI agents and should not be interpreted as a recommendation to deploy autonomous agents on physical laboratory equipment without appropriate safety validation.

\section*{Resources}

Project Page:
\url{https://su-ise-2001.github.io/LABOSBENCH/}

The project website contains benchmark documentation,
simulator demonstrations, task definitions,
evaluation results, and future updates.

\appendix
\section{Benchmark Task Details}
\label{app:task_details}

This appendix provides the complete list of 96 subtasks used in \benchmarkname{}.
Each scientific-instrument simulator is decomposed into a sequence of operation-level subtasks.
For each subtask, we report its name, difficulty level, and natural-language description.
The difficulty levels are assigned according to the type of GUI interaction, the degree of scientific-state interpretation required, and the amount of feedback-driven adjustment involved.

\subsection{Task Construction Principles}
\label{app:task_construction}

The subtasks in \benchmarkname{} are constructed according to scientific workflow stages rather than arbitrary GUI regions.
Each subtask corresponds to an operation-level unit that changes the simulator state, configures an instrument parameter, performs a scientific action, or finalizes an output.
This design preserves the scientific meaning of the original workflow while enabling diagnostic subtask-level evaluation.

\subsection{Complete Subtask List}
\label{app:complete_subtask_list}

The following eight tables summarize the complete subtask lists for all scientific-instrument simulators.
These subtasks are used for the subtask-level evaluation reported in the main paper.

\begin{table*}[t]
\centering
\scriptsize
\setlength{\tabcolsep}{4pt}
\renewcommand{\arraystretch}{1.12}
\caption{XRD subtasks.}
\label{tab:app_xrd_subtasks}
\begin{tabular}{p{3.2cm}p{1.5cm}p{10.0cm}}
\toprule
\textbf{Subtask} & \textbf{Level} & \textbf{Description} \\
\midrule
1. Select Specimen & \textsc{Easy} & Select a specimen from the dropdown menu. \\
2. Load Sample & \textsc{Medium} & Click \texttt{DOORS} to open the chamber, load the sample, and then close the chamber. \\
3. Power Up & \textsc{Medium} & Click \texttt{STANDBY} to power up the instrument and wait until it is ready. \\
4. Set Angles & \textsc{Medium} & Use the angle adjustment buttons to set the start and end angles. \\
5. Set Step Size & \textsc{Medium} & Select the step size from the dropdown menu. \\
6. Set Scan Rate & \textsc{Medium} & Select the scan rate from the dropdown menu. \\
7. Run Scan & \textsc{Easy} & Click \texttt{START SCAN} to begin scanning. \\
8. Save Result & \textsc{Easy} & Click \texttt{SAVE DIFFRACTOGRAM} to save the scan result. \\
\bottomrule
\end{tabular}
\end{table*}

\begin{table*}[t]
\centering
\scriptsize
\setlength{\tabcolsep}{4pt}
\renewcommand{\arraystretch}{1.12}
\caption{FIB subtasks.}
\label{tab:app_fib_subtasks}
\begin{tabular}{p{3.2cm}p{1.5cm}p{10.0cm}}
\toprule
\textbf{Subtask} & \textbf{Level} & \textbf{Description} \\
\midrule
1. Vent Chamber & \textsc{Easy} & Click \texttt{VENT} once and wait for the chamber animation to complete. \\
2. Pump Down & \textsc{Easy} & Click \texttt{PUMP} once and wait for the vacuum process to complete. \\
3. Select Sample & \textsc{Easy} & Select Si Wafer from the SAMPLE dropdown menu. \\
4. E-beam On & \textsc{Easy} & Click the electron-beam \texttt{HT} switch. \\
5. E-beam Live Focus & \textsc{Medium} & Focus the electron-beam image. \\
6. WD 7\,mm & \textsc{Medium} & Adjust the WD slider to approximately 7\,mm. \\
7. Tilt 10$^\circ$ & \textsc{Medium} & Set the stage tilt to approximately 10$^\circ$. \\
8. Stage Z Center & \textsc{Medium} & Move along the Z axis to a suitable height. \\
9. Ion Beam Live Center & \textsc{Medium} & Center the target region in the ion-beam field of view. \\
10. First Rect Start & \textsc{Easy} & Click \texttt{START} once on the correctly selected pattern. \\
11. Delete Pattern & \textsc{Easy} & Click \texttt{DELETE PATTERN} once to remove the current pattern. \\
12. Second Rect Start & \textsc{Medium} & Select the second rectangle and click \texttt{START}. \\
13. Beam Current 10\,pA & \textsc{Medium} & Set the beam current and use \texttt{DELETE PATTERN}. \\
14. Pt Needle In & \textsc{Easy} & Click to extend the Pt needle. \\
15. Pt Deposition Start & \textsc{Hard} & Select the Pt Deposition pattern, place it at the correct position, and click \texttt{START}. \\
16. Ion Snapshot 5000$\times$ & \textsc{Medium} & Set the magnification to approximately 5000$\times$ and capture an ion-beam snapshot. \\
17. Cross Section Cut Start & \textsc{Hard} & Select the cross-section cutting pattern and click in the correct region. \\
18. Cleaning Section Start & \textsc{Hard} & Select the cutting pattern, set the beam current to 0.1\,nA, drag the pattern to the target region, and click \texttt{START}. \\
19. Tilt 0$^\circ$ & \textsc{Easy} & Check Cross Section for imaging and set the stage tilt to 0$^\circ$. \\
20. Task Complete & \textsc{Easy} & Click \texttt{SAVE} to store the cut image. \\
\bottomrule
\end{tabular}
\end{table*}

\begin{table*}[t]
\centering
\scriptsize
\setlength{\tabcolsep}{4pt}
\renewcommand{\arraystretch}{1.12}
\caption{LFM subtasks.}
\label{tab:app_lfm_subtasks}
\begin{tabular}{p{3.2cm}p{1.5cm}p{10.0cm}}
\toprule
\textbf{Subtask} & \textbf{Level} & \textbf{Description} \\
\midrule
1. Drag Sample & \textsc{Easy} & Drag the H\&E-stained kidney specimen onto the microscope stage. \\
2. Select Brightfield & \textsc{Easy} & Check the \texttt{SETUP \& BRIGHTFIELD} mode. \\
3. Select Halogen & \textsc{Easy} & Click \texttt{HALOGEN LAMP} to select the halogen lamp. \\
4. Select 10$\times$ Focus & \textsc{Medium} & Select the 10$\times$ objective and use the \texttt{FOCUS} slider. \\
5. Field Diaphragm Close & \textsc{Easy} & Push the slider fully right to close the field diaphragm. \\
6. Field Diaphragm Focus & \textsc{Hard} & Adjust the slider so that the diaphragm edge is sharply focused. \\
7. Field Diaphragm Center & \textsc{Hard} & Use the up, down, left, and right buttons to center the diaphragm. \\
8. Field Diaphragm Open & \textsc{Easy} & Push the slider left to open the diaphragm until the blades just exceed the field of view. \\
9. Aperture Diaphragm & \textsc{Hard} & Click to remove the right eyepiece, adjust the aperture diaphragm, and reinstall the eyepiece. \\
10. White Balance & \textsc{Hard} & Click the target, move to a blank area, click to set white balance, and then move back. \\
11. Exposure Adjust & \textsc{Medium} & Adjust the exposure slider for optimal exposure and click \texttt{LUT}. \\
12. BF Capture & \textsc{Easy} & Click \texttt{LUT} to check the dynamic range and then click \texttt{CAPTURE}. \\
\bottomrule
\end{tabular}
\end{table*}

\begin{table*}[t]
\centering
\scriptsize
\setlength{\tabcolsep}{4pt}
\renewcommand{\arraystretch}{1.12}
\caption{SEM subtasks.}
\label{tab:app_sem_subtasks}
\begin{tabular}{p{3.2cm}p{1.5cm}p{10.0cm}}
\toprule
\textbf{Subtask} & \textbf{Level} & \textbf{Description} \\
\midrule
1. Vent Chamber & \textsc{Easy} & Click \texttt{VENT} to release vacuum and allow chamber access. \\
2. Open Chamber & \textsc{Easy} & Open the chamber door to load the sample. \\
3. Close Chamber & \textsc{Medium} & Close the chamber door after loading. \\
4. Evacuate Chamber & \textsc{Easy} & Click \texttt{PUMP} to evacuate the chamber. \\
5. Select Sample & \textsc{Easy} & Select the specimen from the SAMPLE dropdown menu. \\
6. Turn On HT & \textsc{Medium} & Click the \texttt{HT} button to turn on the electron beam. \\
7. Set Acc Voltage & \textsc{Hard} & Set the acceleration voltage from the dropdown menu. \\
8. Set Contrast & \textsc{Hard} & Adjust the contrast slider for optimal image quality. \\
9. Adjust Clarity & \textsc{Hard} & Use the focus control to achieve a sharp image. \\
10. Start Scan & \textsc{Easy} & Click \texttt{START SCAN} to begin acquisition. \\
11. Save Image & \textsc{Easy} & Click \texttt{SAVE} to store the acquired image. \\
12. Mosaic ROI Region Save & \textsc{Hard} & In the 2$\times$2 mosaic view, pan or double-click to adjust the field of view, use the MAGNIFICATION slider to zoom into vertically aligned parallel tubular or stripe-like structures, center and sharpen the region, and click \texttt{SAVE IMAGE}. \\
\bottomrule
\end{tabular}
\end{table*}

\begin{table*}[t]
\centering
\scriptsize
\setlength{\tabcolsep}{4pt}
\renewcommand{\arraystretch}{1.12}
\caption{EDS subtasks.}
\label{tab:app_eds_subtasks}
\begin{tabular}{p{3.2cm}p{1.5cm}p{10.0cm}}
\toprule
\textbf{Subtask} & \textbf{Level} & \textbf{Description} \\
\midrule
1. Select Point Mode & \textsc{Easy} & Click \texttt{Point} to select point-analysis mode. \\
2. Pick Sample Point & \textsc{Easy} & Mark the sampling point for subsequent analysis. \\
3. Open Label & \textsc{Easy} & Open the label of the sampling point or spectrum. \\
4. Open Semi Quant & \textsc{Medium} & Open the semi-quantitative analysis window. \\
5. Open Maps & \textsc{Easy} & Open the element distribution map. \\
6. Toggle Si Mapping & \textsc{Easy} & Turn on or off the silicon element distribution map layer. \\
7. Open Composite & \textsc{Easy} & Open the composite diagram. \\
8. Return to Main & \textsc{Easy} & Return to the main control panel. \\
\bottomrule
\end{tabular}
\end{table*}

\begin{table*}[t]
\centering
\scriptsize
\setlength{\tabcolsep}{4pt}
\renewcommand{\arraystretch}{1.12}
\caption{SPM subtasks.}
\label{tab:app_spm_subtasks}
\begin{tabular}{p{3.2cm}p{1.5cm}p{10.0cm}}
\toprule
\textbf{Subtask} & \textbf{Level} & \textbf{Description} \\
\midrule
1. Select Tapping Mode & \textsc{Easy} & Select TAPPING from the MODE dropdown menu. \\
2. Laser Alignment & \textsc{Hard} & Align the laser spot onto the cantilever using the alignment controls. \\
3. Photodiode Alignment & \textsc{Hard} & Center the laser spot on the photodiode using the photodiode controls. \\
4. Set Target Amplitude & \textsc{Easy} & Set the target amplitude to 500\,mV. \\
5. Set Frequency & \textsc{Medium} & Set the frequency range to 200--500\,kHz. \\
6. Auto Tune & \textsc{Easy} & Click the \texttt{AUTO TUNE} button to initiate automated tuning. \\
7. Set Scan Size & \textsc{Easy} & Select the desired scan size from the SCAN SIZE menu. \\
8. Set Integral Gain & \textsc{Medium} & Set the integral gain value for the feedback loop. \\
9. Set Scan Rate & \textsc{Easy} & Select the scan rate from the SCAN RATE menu. \\
10. Set Set Point & \textsc{Medium} & Adjust the set point slider to control the imaging force. \\
11. Motor Approach & \textsc{Medium} & Use the MOTOR slider to bring the probe toward the sample surface. \\
12. Engage & \textsc{Easy} & Click the \texttt{ENGAGE} button to begin the automated approach. \\
13. Scan & \textsc{Easy} & Click the \texttt{SCAN} button to initiate the topographic scan. \\
14. Save & \textsc{Easy} & Click the \texttt{SAVE} button to export the acquired image. \\
\bottomrule
\end{tabular}
\end{table*}

\begin{table*}[t]
\centering
\scriptsize
\setlength{\tabcolsep}{4pt}
\renewcommand{\arraystretch}{1.12}
\caption{APT subtasks.}
\label{tab:app_apt_subtasks}
\begin{tabular}{p{3.2cm}p{1.5cm}p{10.0cm}}
\toprule
\textbf{Subtask} & \textbf{Level} & \textbf{Description} \\
\midrule
1. Select Sample & \textsc{Easy} & Select a sample from the Sample menu. \\
2. Align Sample & \textsc{Hard} & Click \texttt{ALIGN SAMPLE} and wait for the alignment process to finish. \\
3. Set Temperature & \textsc{Medium} & Set the specimen temperature. \\
4. Set Detection & \textsc{Medium} & Set the detection rate. \\
5. Set Frequency & \textsc{Medium} & Set the voltage pulse frequency. \\
6. Set Pulse & \textsc{Medium} & Set the voltage pulse fraction. \\
7. Start Experiment & \textsc{Hard} & Click \texttt{START} to run the experiment. \\
8. Wait Completion & \textsc{Hard} & Wait for the experiment to finish and click again if needed. \\
9. Start Reconstruction & \textsc{Hard} & Click \texttt{RECONSTRUCT} to enter the reconstruction view. \\
10. Set ICF & \textsc{Easy} & Choose an ICF parameter. \\
11. Set K-Factor & \textsc{Easy} & Choose a K-factor parameter. \\
12. Finish Experiment & \textsc{Easy} & Click \texttt{Finish} and then close the popup window. \\
\bottomrule
\end{tabular}
\end{table*}

\begin{table*}[t]
\centering
\scriptsize
\setlength{\tabcolsep}{4pt}
\renewcommand{\arraystretch}{1.12}
\caption{TEM subtasks.}
\label{tab:app_tem_subtasks}
\begin{tabular}{p{3.2cm}p{1.5cm}p{10.0cm}}
\toprule
\textbf{Subtask} & \textbf{Level} & \textbf{Description} \\
\midrule
1. Select Sample & \textsc{Easy} & Select a sample from the SAMPLE dropdown menu. \\
2. Remove Holder & \textsc{Easy} & Click \texttt{REMOVE} to take out the empty holder. \\
3. Pump Airlock & \textsc{Easy} & Click \texttt{PUMP} in the AIRLOCK panel to evacuate the airlock. \\
4. Insert Specimen & \textsc{Easy} & Click \texttt{INSERT} to insert the sample. \\
5. Set Voltage & \textsc{Easy} & Choose an accelerating voltage from the ACC menu. \\
6. Enable Beam & \textsc{Hard} & Turn the beam on and adjust the beam intensity. \\
7. Adjust Magnification & \textsc{Hard} & Select a magnification and adjust image brightness. \\
8. Move Stage & \textsc{Hard} & Move the stage in X, Y, and Z to center the region of interest. \\
9. Focus Objective & \textsc{Hard} & Adjust the OBJECTIVE LENS FOCUS to sharpen the image. \\
10. Acquire Image & \textsc{Easy} & Insert the camera and click \texttt{ACQUIRE} to capture a TEM image. \\
\bottomrule
\end{tabular}
\end{table*}

\subsection{Difficulty Level Definitions}
\label{app:difficulty_levels}

We annotate each subtask with a coarse difficulty level to characterize the interaction complexity.
\textsc{Easy} subtasks typically involve a single explicit GUI action, such as clicking a clearly labeled button or selecting an item from a dropdown menu.
\textsc{Medium} subtasks require a short sequence of actions, parameter adjustment, or waiting for a simulator state transition.
\textsc{Hard} subtasks require feedback-driven control, precise spatial grounding, scientific-state interpretation, or multi-step manipulation under visual feedback.

The difficulty labels are intended for analysis and interpretation rather than for changing the evaluation protocol.
All subtasks are evaluated using the same action interface and success-checking mechanism.

\subsection{Mapping to High-Level Task Categories}
\label{app:category_mapping}

For category-level analysis, each subtask is further mapped to one of five high-level operation categories:
Preparation \& Sample Handling, Activation \& Alignment, Measurement Configuration, Experimental Execution, and Post-processing \& Completion.
The mapping is based on the primary scientific operation required by the subtask rather than the low-level GUI action.
For example, a click on a button may be categorized differently depending on whether it loads a sample, activates a beam, starts a scan, or saves a result.
Table~\ref{tab:app_category_mapping} defines the five high-level categories and lists the operation-level subtasks assigned to each category.

\begin{table*}[t]
\centering
\scriptsize
\setlength{\tabcolsep}{2.8pt}
\renewcommand{\arraystretch}{1.08}
\caption{Definitions of the five high-level task categories and their assigned operation-level subtasks.}
\label{tab:app_category_mapping}
\begin{tabular}{
>{\raggedright\arraybackslash}p{2.9cm}
>{\raggedright\arraybackslash}p{4.0cm}
>{\raggedright\arraybackslash}p{8.4cm}
}
\toprule
\textbf{Task Category} & \textbf{Definition} & \textbf{Assigned Subtasks} \\
\midrule

Preparation \& Sample Handling
&
Operations that prepare the simulator, select the target object, or load the required sample before measurement or manipulation.
&
SEM-S1 VentChamber; SEM-S2 OpenChamber; SEM-S3 CloseChamber; SEM-S4 EvacuateChamber; SEM-S5 SelectSample;
TEM-T1 Select sample; TEM-T2 Remove empty holder; TEM-T3 Airlock pump; TEM-T4 Insert specimen;
XRD-S1 SelectSpecimen; XRD-S2 Doors;
LFM-L1 DragSample;
FIB-F1 VentChamber; FIB-F2 PumpDown; FIB-F3 SelectSample;
APT-S1 SelectSample;
EDS-S2 PickSamplePoint.
\\

\midrule

Activation \& Alignment
&
Operations that activate instrument components or align the system into a valid operating state.
&
SEM-S6 TurnOnHT;
SPM-S1 SelectTappingMode; SPM-S2 LaserAlignment; SPM-S3 PhotodiodeAlignment; SPM-S11 MotorApproach; SPM-S12 Engage;
TEM-T5 Select acceleration voltage; TEM-T6 Beam on and filament current;
XRD-S3 PowerUp;
LFM-L2 SelectBrightfield; LFM-L3 SelectHalogen; LFM-L4 Select10xFocus; LFM-L5 FieldDiaphragmClose; LFM-L6 FieldDiaphragmFocus; LFM-L7 FieldDiaphragmCenter; LFM-L8 FieldDiaphragmOpen; LFM-L9 ApertureDiaphragm;
FIB-F4 EbeamOn; FIB-F5 EbeamLiveFocus; FIB-F8 StageZCenter; FIB-F9 IonBeamLiveCenter;
APT-S2 AlignSample.
\\

\midrule

Measurement Configuration
&
Operations that configure parameters before acquisition, scanning, fabrication, or reconstruction.
&
SEM-S7 SetAccVoltage; SEM-S8 SetContrast; SEM-S9 AdjustClarity;
SPM-S4 SetTargetAmplitude; SPM-S5 SetFrequency; SPM-S6 AutoTune; SPM-S7 SetScanSize; SPM-S8 SetIntegralGain; SPM-S9 SetScanRate; SPM-S10 SetSetPoint;
TEM-T7 Magnification and brightness; TEM-T8 Stage position centering; TEM-T9 Objective lens focus;
XRD-S4 SetAngles; XRD-S5 SetStepSize; XRD-S6 SetScanRate;
LFM-L10 WhiteBalance; LFM-L11 ExposureAdjust;
FIB-F6 WD7mm; FIB-F7 Tilt10deg; FIB-F13 BeamCurrent10pA; FIB-F16 IonSnapshot5000x;
APT-S3 SelectTemperature; APT-S4 SelectDetectionRate; APT-S5 SelectPulseFreq; APT-S6 SelectPulseEnergy;
EDS-S1 SelectPointMode.
\\

\midrule

Experimental Execution
&
Operations that start, monitor, or complete the main scientific action after the setup is ready.
&
SEM-S10 StartScan;
SPM-S13 Scan;
TEM-T10 Camera insert and acquire;
XRD-S7 RunScan;
LFM-L12 BFCapture;
FIB-F10 FirstRectStart; FIB-F11 DeletePattern; FIB-F12 SecondRectStart; FIB-F14 PtNeedleIn; FIB-F15 PtDepositionStart; FIB-F17 CrossSectionCutStart; FIB-F18 CleaningCrossSectionStart;
APT-S7 StartExperiment; APT-S8 WaitCompletion.
\\

\midrule

Post-processing \& Completion
&
Operations that inspect, save, export, reconstruct, or finalize experimental outputs.
&
SEM-S11 SaveImage; SEM-S12 MosaicRoiRegionSave;
SPM-S14 Save;
XRD-S8 SaveResult;
FIB-F19 Tilt0deg; FIB-F20 TaskComplete;
APT-S9 Reconstruct; APT-S10 SetICF; APT-S11 SetKFactor; APT-S12 Finish;
EDS-S3 OpenLabel; EDS-S4 OpenSemiQuant; EDS-S5 OpenMaps; EDS-S6 ToggleSiMapping; EDS-S7 OpenComposite; EDS-S8 ReturnToMain.
\\

\bottomrule
\end{tabular}
\end{table*}

\section{Evaluation Protocol and Implementation Details}
\label{app:protocol}

This appendix provides additional details about the evaluation protocol and implementation of \benchmarkname{}.
The goal is to make the browser-based execution process, action interface, subtask initialization, step budget, and logging format explicit.
All models and agentic frameworks are evaluated through the same interaction protocol unless otherwise specified.

\subsection{Observation and Action Space}
\label{app:action_space}

At each interaction step, the agent receives a natural-language task instruction and the current screenshot of the simulator interface.
The agent then outputs one executable GUI action.
We use a unified action interface across all simulators so that the same evaluation runner can be applied to different scientific-instrument workflows.

The action space includes clicking, double-clicking, dragging, typing, selecting dropdown options, pressing keys, scrolling, and waiting for simulator state transitions.
Click and double-click actions are used for buttons, panels, and visual target regions; drag actions support sliders, stage movement, and pattern placement; type and select actions handle numeric inputs and menus; wait actions allow instrument animations, scans, reconstructions, or state transitions to complete.

For coordinate-based actions, the runner maps the agent-produced coordinate to the current browser viewport.
When an action targets a standard HTML element, the runner uses DOM-level event dispatch when possible.
For canvas-like or JavaScript-rendered controls, it falls back to mouse events.
This hybrid execution strategy allows \benchmarkname{} to support both conventional web widgets and simulator-specific interactive controls.

\subsection{Interaction Loop}
\label{app:interaction_loop}

Each evaluation episode follows a closed-loop observe--act--execute protocol.
At the beginning of an episode, the runner launches the target simulator page, applies the required initial state, and provides the agent with the task instruction and the initial screenshot.
At each step, the agent predicts the next GUI action based on the instruction and the current observation.
The runner parses the action, executes it in the browser, waits for the simulator state to update, captures a new screenshot, and queries the in-page benchmark state.

The loop terminates when the target subtask or workflow is successfully completed, when the step budget is exhausted, or when the runner detects an unrecoverable execution error.
Invalid or unparsable actions are recorded in the episode log and counted as failed interaction attempts.
This protocol allows the benchmark to capture both successful task completion and intermediate failure behavior, such as repeated clicks, wrong coordinate selection, ineffective waiting, or incorrect parameter manipulation.

\subsection{Subtask Initialization}
\label{app:subtask_initialization}

For subtask-level evaluation, each simulator is initialized to a canonical state immediately before the target subtask.
This initialization isolates the local capability required by the target operation and avoids conflating subtask performance with failures accumulated in earlier workflow stages.
For example, when evaluating a parameter-setting subtask, the simulator is first placed into the state where the relevant control is available and all prerequisite subtasks have already been completed.

When a simulator supports programmatic state control, we use instrument-specific fast-forward functions exposed through the browser window.
These functions mark prerequisite subtasks as completed and restore the GUI to the canonical pre-subtask state.
Fast-forwarding is used only for diagnostic subtask-level evaluation.
It is not used in full-episode evaluation, where the agent must complete the entire workflow from the initial simulator state.

This design allows \benchmarkname{} to distinguish local GUI-control ability from long-horizon workflow execution.
An agent that succeeds under subtask initialization but fails in full-episode evaluation may suffer from planning, ordering, memory, or recovery problems.
In contrast, failure under subtask initialization indicates that the agent cannot reliably complete the local scientific-instrument operation itself.

\subsection{Step Budget and Termination}
\label{app:step_budget}

For subtask-level evaluation, each model is allowed up to 50 interaction steps per subtask.
A subtask episode is marked as successful if the corresponding success condition is satisfied within the step budget.
Otherwise, the episode is marked as failed after the budget is exhausted.

For full-episode evaluation, the agent starts from the initial simulator state and must complete all required subtasks in order.
A full episode is successful only if all required workflow stages are completed within the specified workflow-level step budget.
This setting evaluates long-horizon execution and captures error accumulation across the entire scientific workflow.

An episode may terminate under three conditions:
(i) all target success conditions are satisfied,
(ii) the maximum number of allowed interaction steps is reached, or
(iii) the runner encounters an unrecoverable execution error.
All termination causes are recorded in the episode log.

\subsection{Logging Format}
\label{app:logging_format}

After each evaluation, the runner exports a structured JSON log for each simulator and model.
The log records the number of runs per subtask, subtask-level outcomes, run-level episode metadata, diagnostic grounding metrics, and overall aggregated statistics.
This format allows us to compute the reported subtask success rates while preserving additional diagnostic information for failure analysis.

At the top level, each log contains \texttt{num\_runs\_per\_subtask}, a list of \texttt{subtasks}, and an \texttt{overall\_metrics\_summary}.
Each subtask entry stores the subtask identifier, subtask name, number of runs, success count, success rate, averaged diagnostic metrics, and the detailed records of individual runs.
Each run record stores whether the run succeeds, the directory of the corresponding episode trace, and run-level metrics such as actual interaction steps, grounding accuracy, target-subtask attempts, and target-subtask completion.

\begin{table*}[t]
\centering
\scriptsize
\caption{Main fields in the exported evaluation logs.}
\label{tab:app_log_fields}
\begin{tabular}{p{3.0cm}p{3.4cm}p{8.2cm}}
\toprule
\textbf{Level} & \textbf{Field} & \textbf{Description} \\
\midrule
Global & \texttt{num\_runs\_per\_subtask} & Number of repeated runs used for each subtask. \\
Global & \texttt{subtasks} & List of subtask-level evaluation records. \\
Global & \texttt{overall\_metrics\_summary} & Aggregated diagnostic metrics across all subtasks in the simulator. \\
\midrule
Subtask & \texttt{subtask\_id} & Identifier of the target subtask, such as \texttt{S1}. \\
Subtask & \texttt{subtask\_name} & Name of the target subtask. \\
Subtask & \texttt{num\_runs} & Number of runs available for the subtask. \\
Subtask & \texttt{success\_count} & Number of successful runs for the subtask. \\
Subtask & \texttt{success\_rate} & Success rate of the subtask over repeated runs. \\
Subtask & \texttt{metrics\_summary} & Averaged diagnostic metrics over runs, such as actual steps, grounding accuracy, and target-subtask attempts. \\
Subtask & \texttt{runs} & Detailed records for individual runs. \\
\midrule
Run & \texttt{run} & Index of the repeated run. \\
Run & \texttt{success} & Whether the run is judged successful by the evaluation script. \\
Run & \texttt{episode\_dir} & Directory storing the episode trace and screenshots. Paths are anonymized in released logs. \\
Run & \texttt{metrics} & Run-level diagnostic metrics. \\
\midrule
Metric & \texttt{actual\_steps} & Number of interaction steps used by the agent. \\
Metric & \texttt{target\_subtask\_attempts} & Number of detected attempts related to the target subtask. \\
Metric & \texttt{widget\_grounding\_accuracy} & Whether the agent correctly grounds the target GUI widget when applicable. \\
Metric & \texttt{text\_grounding\_accuracy} & Whether the agent correctly grounds the relevant text or label when applicable. \\
Metric & \texttt{state\_grounding\_accuracy} & Whether the agent correctly reaches or recognizes the required simulator state when applicable. \\
Metric & \texttt{target\_subtask\_success} & Whether the target-subtask event is detected in the in-page log. \\
\bottomrule
\end{tabular}
\end{table*}

The official subtask-level score is computed from \texttt{success\_count} and \texttt{success\_rate}.
Other fields, such as \texttt{actual\_steps}, \texttt{target\_subtask\_attempts}, \texttt{widget\_grounding\_accuracy}, \texttt{text\_grounding\_accuracy}, \texttt{state\_grounding\_accuracy}, and \texttt{target\_subtask\_success}, are used as diagnostic signals rather than as the primary score.
This distinction allows the benchmark to report final task success while still analyzing intermediate grounding behavior and failure modes.
For anonymity, machine-specific absolute paths in \texttt{episode\_dir} are removed or anonymized before release.

\subsection{Implementation Notes}
\label{app:implementation_notes}

The evaluation runner is implemented as a lightweight browser-based pipeline.
It launches local simulator pages, captures screenshots, dispatches GUI actions, and exports structured logs after each episode.
This design avoids full operating-system virtualization while retaining executable interaction with realistic scientific-instrument interfaces.
Because the simulator state is logged at each step, the same traces can be used for quantitative scoring, failure localization, and qualitative case analysis.

\section{Subtask-level Results}
\label{app:subtask_results}

This appendix reports the subtask-level success rates of all evaluated models on the \benchmarkname{} benchmark.
Tables~\ref{tab:app_sem_subtask_results}--\ref{tab:app_eds_subtask_results} provide the complete results grouped by scientific-instrument simulator.
Each row represents an operation-level subtask, and all scores are reported in the range $[0,1]$.

\begin{table*}[t]
\centering
\scriptsize
\setlength{\tabcolsep}{1.6pt}
\renewcommand{\arraystretch}{1.08}
\caption{Subtask-level success rates on SEM.}
\label{tab:app_sem_subtask_results}
\resizebox{\textwidth}{!}{%
\begin{tabular}{llcccccccccccc}
\toprule
\textbf{ID} & \textbf{Subtask} & \textbf{Qwen3VL-32B} & \textbf{EvoCUA} & \textbf{Sonnet-4.5} & \textbf{Kimi} & \textbf{Seed} & \textbf{GPT-5.5} & \textbf{Opus-4.5} & \textbf{UI-TARS} & \textbf{GUI-Owl} & \textbf{GTA1} & \textbf{VLAA} & \textbf{Hippo} \\
\midrule
S1 & VentChamber & 1.00 & 1.00 & 1.00 & 0.50 & 1.00 & 1.00 & 1.00 & 1.00 & 1.00 & 1.00 & 1.00 & 1.00 \\
S2 & OpenChamber & 1.00 & 1.00 & 1.00 & 1.00 & 1.00 & 1.00 & 1.00 & 1.00 & 1.00 & 1.00 & 1.00 & 1.00 \\
S3 & CloseChamber & 1.00 & 1.00 & 1.00 & 1.00 & 1.00 & 1.00 & 1.00 & 0.00 & 1.00 & 1.00 & 1.00 & 1.00 \\
S4 & EvacuateChamber & 1.00 & 1.00 & 1.00 & 1.00 & 1.00 & 1.00 & 1.00 & 1.00 & 1.00 & 1.00 & 1.00 & 1.00 \\
S5 & SelectSample & 0.50 & 0.00 & 0.00 & 0.00 & 1.00 & 0.00 & 0.00 & 1.00 & 0.50 & 0.50 & 0.00 & 0.50 \\
S6 & TurnOnHT & 1.00 & 1.00 & 1.00 & 1.00 & 1.00 & 1.00 & 1.00 & 1.00 & 1.00 & 1.00 & 0.00 & 1.00 \\
S7 & SetAccVoltage & 1.00 & 1.00 & 1.00 & 0.50 & 1.00 & 1.00 & 1.00 & 0.00 & 1.00 & 1.00 & 0.00 & 1.00 \\
S8 & SetContrast & 0.50 & 1.00 & 1.00 & 0.00 & 0.00 & 1.00 & 1.00 & 0.00 & 0.50 & 1.00 & 0.00 & 0.50 \\
S9 & AdjustClarity & 1.00 & 1.00 & 1.00 & 0.50 & 0.50 & 1.00 & 1.00 & 0.00 & 1.00 & 1.00 & 0.00 & 1.00 \\
S10 & StartScan & 1.00 & 1.00 & 1.00 & 1.00 & 1.00 & 1.00 & 1.00 & 1.00 & 1.00 & 1.00 & 0.00 & 1.00 \\
S11 & SaveImage & 1.00 & 1.00 & 1.00 & 1.00 & 1.00 & 1.00 & 1.00 & 0.00 & 1.00 & 1.00 & 0.00 & 1.00 \\
S12 & MosaicRoiRegionSave & 0.00 & 0.00 & 0.00 & 0.00 & 0.00 & 0.00 & 0.00 & 0.00 & 0.00 & 0.00 & 0.00 & 0.00 \\
\bottomrule
\end{tabular}%
}
\end{table*}

\begin{table*}[t]
\centering
\scriptsize
\setlength{\tabcolsep}{1.6pt}
\renewcommand{\arraystretch}{1.08}
\caption{Subtask-level success rates on SPM.}
\label{tab:app_spm_subtask_results}
\resizebox{\textwidth}{!}{%
\begin{tabular}{llcccccccccccc}
\toprule
\textbf{ID} & \textbf{Subtask} & \textbf{Qwen3VL-32B} & \textbf{EvoCUA} & \textbf{Sonnet-4.5} & \textbf{Kimi} & \textbf{Seed} & \textbf{GPT-5.5} & \textbf{Opus-4.5} & \textbf{UI-TARS} & \textbf{GUI-Owl} & \textbf{GTA1} & \textbf{VLAA} & \textbf{Hippo} \\
\midrule
S1 & SelectTappingMode & 1.00 & 1.00 & 1.00 & 1.00 & 1.00 & 1.00 & 1.00 & 1.00 & 1.00 & 1.00 & 1.00 & 1.00 \\
S2 & LaserAlignment & 0.00 & 0.00 & 0.00 & 0.00 & 0.00 & 0.00 & 0.00 & 0.00 & 0.00 & 0.50 & 0.00 & 0.00 \\
S3 & PhotodiodeAlignment & 0.00 & 0.00 & 0.00 & 0.00 & 0.00 & 0.00 & 0.00 & 0.00 & 0.00 & 0.00 & 0.00 & 0.00 \\
S4 & SetTargetAmplitude & 0.50 & 0.00 & 1.00 & 0.00 & 1.00 & 1.00 & 1.00 & 1.00 & 1.00 & 1.00 & 0.00 & 1.00 \\
S5 & SetFrequency & 0.50 & 0.00 & 1.00 & 0.00 & 1.00 & 1.00 & 1.00 & 1.00 & 0.50 & 1.00 & 0.00 & 0.50 \\
S6 & AutoTune & 1.00 & 1.00 & 1.00 & 1.00 & 1.00 & 1.00 & 1.00 & 1.00 & 1.00 & 1.00 & 1.00 & 1.00 \\
S7 & SetScanSize & 1.00 & 1.00 & 1.00 & 1.00 & 1.00 & 1.00 & 1.00 & 1.00 & 1.00 & 1.00 & 0.00 & 1.00 \\
S8 & SetIntegralGain & 0.50 & 1.00 & 0.00 & 1.00 & 1.00 & 1.00 & 0.00 & 1.00 & 0.50 & 1.00 & 0.00 & 0.50 \\
S9 & SetScanRate & 1.00 & 0.00 & 1.00 & 0.50 & 1.00 & 1.00 & 1.00 & 1.00 & 1.00 & 1.00 & 1.00 & 1.00 \\
S10 & SetSetPoint & 0.50 & 1.00 & 0.00 & 0.00 & 0.00 & 1.00 & 0.00 & 0.50 & 0.50 & 0.50 & 0.00 & 0.50 \\
S11 & MotorApproach & 0.00 & 0.00 & 1.00 & 0.00 & 1.00 & 0.00 & 0.00 & 0.00 & 0.50 & 0.50 & 0.00 & 0.50 \\
S12 & Engage & 1.00 & 1.00 & 1.00 & 0.00 & 1.00 & 1.00 & 1.00 & 1.00 & 1.00 & 1.00 & 1.00 & 1.00 \\
S13 & Scan & 1.00 & 1.00 & 1.00 & 0.00 & 1.00 & 1.00 & 1.00 & 1.00 & 1.00 & 1.00 & 1.00 & 1.00 \\
S14 & Save & 1.00 & 1.00 & 1.00 & 0.50 & 1.00 & 1.00 & 1.00 & 1.00 & 1.00 & 1.00 & 1.00 & 1.00 \\
\bottomrule
\end{tabular}%
}
\end{table*}

\begin{table*}[t]
\centering
\scriptsize
\setlength{\tabcolsep}{1.6pt}
\renewcommand{\arraystretch}{1.08}
\caption{Subtask-level success rates on TEM.}
\label{tab:app_tem_subtask_results}
\resizebox{\textwidth}{!}{%
\begin{tabular}{llcccccccccccc}
\toprule
\textbf{ID} & \textbf{Subtask} & \textbf{Qwen3VL-32B} & \textbf{EvoCUA} & \textbf{Sonnet-4.5} & \textbf{Kimi} & \textbf{Seed} & \textbf{GPT-5.5} & \textbf{Opus-4.5} & \textbf{UI-TARS} & \textbf{GUI-Owl} & \textbf{GTA1} & \textbf{VLAA} & \textbf{Hippo} \\
\midrule
T1 & Select sample & 0.50 & 1.00 & 0.00 & 1.00 & 1.00 & 0.00 & 0.00 & 1.00 & 0.50 & 1.00 & 1.00 & 0.50 \\
T2 & Remove empty holder & 1.00 & 1.00 & 1.00 & 1.00 & 1.00 & 1.00 & 1.00 & 1.00 & 1.00 & 1.00 & 1.00 & 1.00 \\
T3 & Airlock pump & 0.50 & 0.00 & 0.00 & 0.00 & 1.00 & 1.00 & 0.00 & 1.00 & 0.50 & 1.00 & 0.00 & 0.50 \\
T4 & Insert specimen & 1.00 & 1.00 & 1.00 & 0.00 & 1.00 & 1.00 & 1.00 & 1.00 & 1.00 & 1.00 & 1.00 & 1.00 \\
T5 & Select acceleration voltage & 0.50 & 1.00 & 1.00 & 1.00 & 1.00 & 0.00 & 1.00 & 1.00 & 0.50 & 1.00 & 0.00 & 0.50 \\
T6 & Beam on and filament current & 0.00 & 0.00 & 0.00 & 1.00 & 0.50 & 1.00 & 0.00 & 0.00 & 0.00 & 0.50 & 0.00 & 0.00 \\
T7 & Magnification and brightness & 1.00 & 1.00 & 1.00 & 0.00 & 0.00 & 1.00 & 1.00 & 0.00 & 1.00 & 1.00 & 1.00 & 1.00 \\
T8 & Stage position centering & 0.00 & 0.00 & 0.00 & 0.00 & 0.00 & 0.00 & 0.00 & 0.00 & 0.00 & 0.50 & 0.00 & 0.00 \\
T9 & Objective lens focus & 0.00 & 0.00 & 0.00 & 0.00 & 0.00 & 0.00 & 0.00 & 0.00 & 0.00 & 0.50 & 0.00 & 0.00 \\
T10 & Camera insert and acquire & 0.50 & 0.00 & 0.00 & 1.00 & 1.00 & 0.00 & 0.00 & 1.00 & 0.50 & 1.00 & 0.00 & 0.50 \\
\bottomrule
\end{tabular}%
}
\end{table*}

\begin{table*}[t]
\centering
\scriptsize
\setlength{\tabcolsep}{1.6pt}
\renewcommand{\arraystretch}{1.08}
\caption{Subtask-level success rates on XRD.}
\label{tab:app_xrd_subtask_results}
\resizebox{\textwidth}{!}{%
\begin{tabular}{llcccccccccccc}
\toprule
\textbf{ID} & \textbf{Subtask} & \textbf{Qwen3VL-32B} & \textbf{EvoCUA} & \textbf{Sonnet-4.5} & \textbf{Kimi} & \textbf{Seed} & \textbf{GPT-5.5} & \textbf{Opus-4.5} & \textbf{UI-TARS} & \textbf{GUI-Owl} & \textbf{GTA1} & \textbf{VLAA} & \textbf{Hippo} \\
\midrule
S1 & SelectSpecimen & 0.50 & 1.00 & 1.00 & 1.00 & 1.00 & 0.00 & 1.00 & 1.00 & 0.50 & 1.00 & 1.00 & 0.50 \\
S2 & Doors & 1.00 & 1.00 & 0.00 & 1.00 & 1.00 & 1.00 & 1.00 & 1.00 & 1.00 & 1.00 & 1.00 & 1.00 \\
S3 & PowerUp & 1.00 & 1.00 & 1.00 & 1.00 & 1.00 & 1.00 & 1.00 & 0.50 & 1.00 & 1.00 & 1.00 & 1.00 \\
S4 & SetAngles & 0.50 & 0.00 & 0.00 & 1.00 & 1.00 & 0.00 & 0.00 & 1.00 & 0.50 & 0.50 & 0.00 & 0.50 \\
S5 & SetStepSize & 1.00 & 1.00 & 1.00 & 1.00 & 1.00 & 1.00 & 1.00 & 1.00 & 1.00 & 1.00 & 0.00 & 1.00 \\
S6 & SetScanRate & 0.50 & 1.00 & 1.00 & 0.00 & 0.00 & 1.00 & 1.00 & 0.00 & 0.50 & 1.00 & 1.00 & 0.50 \\
S7 & RunScan & 0.50 & 0.00 & 0.00 & 1.00 & 1.00 & 0.00 & 0.00 & 1.00 & 0.50 & 0.50 & 0.00 & 0.50 \\
S8 & SaveResult & 1.00 & 1.00 & 1.00 & 1.00 & 1.00 & 1.00 & 1.00 & 1.00 & 1.00 & 1.00 & 1.00 & 1.00 \\
\bottomrule
\end{tabular}%
}
\end{table*}

\begin{table*}[t]
\centering
\scriptsize
\setlength{\tabcolsep}{1.6pt}
\renewcommand{\arraystretch}{1.08}
\caption{Subtask-level success rates on LFM.}
\label{tab:app_lfm_subtask_results}
\resizebox{\textwidth}{!}{%
\begin{tabular}{llcccccccccccc}
\toprule
\textbf{ID} & \textbf{Subtask} & \textbf{Qwen3VL-32B} & \textbf{EvoCUA} & \textbf{Sonnet-4.5} & \textbf{Kimi} & \textbf{Seed} & \textbf{GPT-5.5} & \textbf{Opus-4.5} & \textbf{UI-TARS} & \textbf{GUI-Owl} & \textbf{GTA1} & \textbf{VLAA} & \textbf{Hippo} \\
\midrule
L1 & DragSample & 1.00 & 1.00 & 1.00 & 1.00 & 1.00 & 1.00 & 1.00 & 1.00 & 0.50 & 1.00 & 1.00 & 1.00 \\
L2 & SelectBrightfield & 1.00 & 1.00 & 1.00 & 1.00 & 1.00 & 1.00 & 1.00 & 1.00 & 0.50 & 1.00 & 1.00 & 1.00 \\
L3 & SelectHalogen & 0.00 & 0.00 & 0.00 & 1.00 & 1.00 & 0.00 & 0.00 & 0.50 & 0.00 & 0.50 & 0.00 & 0.50 \\
L4 & Select10xFocus & 0.00 & 0.00 & 0.00 & 0.00 & 1.00 & 0.00 & 0.00 & 0.00 & 0.00 & 0.50 & 0.00 & 0.00 \\
L5 & FieldDiaphragmClose & 1.00 & 1.00 & 1.00 & 1.00 & 1.00 & 1.00 & 1.00 & 1.00 & 0.50 & 1.00 & 0.00 & 1.00 \\
L6 & FieldDiaphragmFocus & 0.00 & 0.00 & 0.00 & 0.00 & 0.00 & 0.00 & 0.00 & 0.00 & 0.00 & 0.00 & 0.00 & 0.00 \\
L7 & FieldDiaphragmCenter & 0.00 & 0.00 & 0.00 & 0.00 & 0.00 & 0.00 & 0.00 & 0.00 & 0.00 & 0.50 & 0.00 & 0.00 \\
L8 & FieldDiaphragmOpen & 1.00 & 1.00 & 1.00 & 1.00 & 1.00 & 1.00 & 1.00 & 1.00 & 0.50 & 1.00 & 0.00 & 1.00 \\
L9 & ApertureDiaphragm & 0.00 & 0.00 & 0.00 & 0.00 & 0.00 & 0.00 & 0.00 & 0.00 & 0.00 & 0.50 & 0.00 & 0.00 \\
L10 & WhiteBalance & 0.00 & 0.00 & 0.00 & 1.00 & 0.00 & 0.00 & 0.00 & 0.00 & 0.00 & 0.50 & 0.00 & 0.50 \\
L11 & ExposureAdjust & 1.00 & 1.00 & 1.00 & 1.00 & 1.00 & 1.00 & 1.00 & 1.00 & 1.00 & 1.00 & 1.00 & 1.00 \\
L12 & BFCapture & 1.00 & 1.00 & 1.00 & 1.00 & 1.00 & 1.00 & 1.00 & 1.00 & 1.00 & 1.00 & 1.00 & 1.00 \\
\bottomrule
\end{tabular}%
}
\end{table*}

\begin{table*}[t]
\centering
\scriptsize
\setlength{\tabcolsep}{1.6pt}
\renewcommand{\arraystretch}{1.08}
\caption{Subtask-level success rates on FIB.}
\label{tab:app_fib_subtask_results}
\resizebox{\textwidth}{!}{%
\begin{tabular}{llcccccccccccc}
\toprule
\textbf{ID} & \textbf{Subtask} & \textbf{Qwen3VL-32B} & \textbf{EvoCUA} & \textbf{Sonnet-4.5} & \textbf{Kimi} & \textbf{Seed} & \textbf{GPT-5.5} & \textbf{Opus-4.5} & \textbf{UI-TARS} & \textbf{GUI-Owl} & \textbf{GTA1} & \textbf{VLAA} & \textbf{Hippo} \\
\midrule
F1 & VentChamber & 0.50 & 1.00 & 1.00 & 1.00 & 1.00 & 1.00 & 1.00 & 1.00 & 1.00 & 1.00 & 1.00 & 1.00 \\
F2 & PumpDown & 1.00 & 1.00 & 1.00 & 0.50 & 1.00 & 1.00 & 1.00 & 1.00 & 1.00 & 1.00 & 1.00 & 1.00 \\
F3 & SelectSample & 1.00 & 1.00 & 1.00 & 1.00 & 1.00 & 1.00 & 1.00 & 1.00 & 1.00 & 1.00 & 1.00 & 1.00 \\
F4 & EbeamOn & 0.50 & 0.00 & 1.00 & 0.50 & 1.00 & 1.00 & 1.00 & 1.00 & 0.50 & 1.00 & 0.00 & 0.50 \\
F5 & EbeamLiveFocus & 0.00 & 0.00 & 0.00 & 0.50 & 0.00 & 1.00 & 0.00 & 0.00 & 0.00 & 0.50 & 0.00 & 0.00 \\
F6 & WD7mm & 0.50 & 0.00 & 1.00 & 0.50 & 0.50 & 1.00 & 1.00 & 1.00 & 0.50 & 1.00 & 0.00 & 0.50 \\
F7 & Tilt10deg & 0.00 & 0.00 & 0.00 & 0.00 & 0.00 & 0.00 & 0.00 & 0.00 & 0.00 & 0.00 & 0.00 & 0.00 \\
F8 & StageZCenter & 0.00 & 0.00 & 0.00 & 0.00 & 0.00 & 0.00 & 0.00 & 0.00 & 0.00 & 0.00 & 0.00 & 0.00 \\
F9 & IonBeamLiveCenter & 0.00 & 0.00 & 0.00 & 0.00 & 0.00 & 0.00 & 0.00 & 0.00 & 0.00 & 0.50 & 0.00 & 0.00 \\
F10 & FirstRectStart & 0.00 & 0.00 & 0.00 & 1.00 & 1.00 & 0.00 & 0.00 & 1.00 & 0.00 & 0.50 & 0.00 & 0.00 \\
F11 & DeletePattern & 0.00 & 0.00 & 0.00 & 0.50 & 0.50 & 1.00 & 0.00 & 1.00 & 0.50 & 0.50 & 0.00 & 0.00 \\
F12 & SecondRectStart & 0.00 & 0.00 & 0.00 & 0.00 & 0.00 & 0.00 & 0.00 & 0.00 & 0.00 & 0.50 & 0.00 & 0.00 \\
F13 & BeamCurrent10pA & 0.50 & 1.00 & 0.00 & 0.50 & 1.00 & 1.00 & 0.00 & 1.00 & 0.50 & 1.00 & 0.00 & 0.50 \\
F14 & PtNeedleIn & 0.50 & 1.00 & 0.00 & 1.00 & 1.00 & 1.00 & 0.00 & 1.00 & 0.50 & 1.00 & 0.00 & 0.50 \\
F15 & PtDepositionStart & 0.00 & 0.00 & 0.00 & 0.00 & 0.00 & 1.00 & 0.00 & 0.00 & 0.00 & 0.50 & 0.00 & 0.00 \\
F16 & IonSnapshot5000x & 0.50 & 1.00 & 0.00 & 1.00 & 1.00 & 1.00 & 0.00 & 1.00 & 0.50 & 1.00 & 0.00 & 0.50 \\
F17 & CrossSectionCutStart & 0.00 & 0.00 & 0.00 & 0.50 & 1.00 & 1.00 & 0.00 & 0.00 & 0.00 & 0.50 & 0.00 & 0.00 \\
F18 & CleaningCrossSectionStart & 0.50 & 1.00 & 0.00 & 1.00 & 1.00 & 1.00 & 0.00 & 1.00 & 0.50 & 1.00 & 0.00 & 0.50 \\
F19 & Tilt0deg & 0.00 & 0.00 & 0.00 & 0.50 & 1.00 & 0.00 & 0.00 & 1.00 & 0.00 & 0.50 & 0.00 & 0.00 \\
F20 & TaskComplete & 0.00 & 0.00 & 0.00 & 0.50 & 1.00 & 0.00 & 0.00 & 1.00 & 0.00 & 0.50 & 0.00 & 0.00 \\
\bottomrule
\end{tabular}%
}
\end{table*}

\begin{table*}[t]
\centering
\scriptsize
\setlength{\tabcolsep}{1.6pt}
\renewcommand{\arraystretch}{1.08}
\caption{Subtask-level success rates on APT.}
\label{tab:app_apt_subtask_results}
\resizebox{\textwidth}{!}{%
\begin{tabular}{llcccccccccccc}
\toprule
\textbf{ID} & \textbf{Subtask} & \textbf{Qwen3VL-32B} & \textbf{EvoCUA} & \textbf{Sonnet-4.5} & \textbf{Kimi} & \textbf{Seed} & \textbf{GPT-5.5} & \textbf{Opus-4.5} & \textbf{UI-TARS} & \textbf{GUI-Owl} & \textbf{GTA1} & \textbf{VLAA} & \textbf{Hippo} \\
\midrule
S1 & SelectSample & 1.00 & 0.50 & 1.00 & 1.00 & 1.00 & 1.00 & 1.00 & 1.00 & 1.00 & 1.00 & 1.00 & 1.00 \\
S2 & AlignSample & 0.50 & 1.00 & 0.00 & 0.00 & 0.50 & 1.00 & 0.00 & 0.00 & 0.50 & 0.50 & 0.00 & 0.50 \\
S3 & SelectTemperature & 0.50 & 1.00 & 0.00 & 1.00 & 1.00 & 1.00 & 0.00 & 1.00 & 0.50 & 1.00 & 0.00 & 0.50 \\
S4 & SelectDetectionRate & 1.00 & 1.00 & 1.00 & 1.00 & 1.00 & 1.00 & 1.00 & 1.00 & 1.00 & 1.00 & 1.00 & 1.00 \\
S5 & SelectPulseFreq & 0.50 & 0.50 & 1.00 & 1.00 & 1.00 & 1.00 & 1.00 & 1.00 & 0.50 & 1.00 & 0.00 & 0.50 \\
S6 & SelectPulseEnergy & 1.00 & 1.00 & 1.00 & 1.00 & 1.00 & 1.00 & 1.00 & 1.00 & 1.00 & 1.00 & 1.00 & 1.00 \\
S7 & StartExperiment & 0.50 & 1.00 & 1.00 & 0.00 & 1.00 & 1.00 & 1.00 & 0.00 & 0.50 & 1.00 & 1.00 & 0.50 \\
S8 & StopExperiment & 1.00 & 1.00 & 1.00 & 0.00 & 1.00 & 1.00 & 1.00 & 0.00 & 1.00 & 1.00 & 1.00 & 0.50 \\
S9 & Reconstruct & 0.50 & 1.00 & 1.00 & 0.00 & 0.00 & 1.00 & 0.00 & 0.00 & 0.50 & 0.50 & 0.00 & 0.50 \\
S10 & SetICF & 0.50 & 0.50 & 0.00 & 1.00 & 1.00 & 1.00 & 0.00 & 1.00 & 0.50 & 0.50 & 0.00 & 0.50 \\
S11 & SetKFactor & 0.50 & 0.00 & 0.00 & 1.00 & 1.00 & 0.50 & 0.00 & 1.00 & 0.50 & 0.50 & 0.00 & 0.00 \\
S12 & Finish & 0.50 & 0.00 & 1.00 & 1.00 & 1.00 & 0.50 & 1.00 & 1.00 & 0.50 & 1.00 & 1.00 & 0.50 \\
\bottomrule
\end{tabular}%
}
\end{table*}

\begin{table*}[t]
\centering
\scriptsize
\setlength{\tabcolsep}{1.6pt}
\renewcommand{\arraystretch}{1.08}
\caption{Subtask-level success rates on EDS.}
\label{tab:app_eds_subtask_results}
\resizebox{\textwidth}{!}{%
\begin{tabular}{llcccccccccccc}
\toprule
\textbf{ID} & \textbf{Subtask} & \textbf{Qwen3VL-32B} & \textbf{EvoCUA} & \textbf{Sonnet-4.5} & \textbf{Kimi} & \textbf{Seed} & \textbf{GPT-5.5} & \textbf{Opus-4.5} & \textbf{UI-TARS} & \textbf{GUI-Owl} & \textbf{GTA1} & \textbf{VLAA} & \textbf{Hippo} \\
\midrule
S1 & SelectPointMode & 0.50 & 1.00 & 1.00 & 0.00 & 1.00 & 1.00 & 1.00 & 1.00 & 1.00 & 1.00 & 1.00 & 1.00 \\
S2 & PickSamplePoint & 0.50 & 1.00 & 1.00 & 1.00 & 0.50 & 1.00 & 1.00 & 0.50 & 1.00 & 0.50 & 0.00 & 1.00 \\
S3 & OpenLabel & 0.50 & 1.00 & 1.00 & 1.00 & 1.00 & 1.00 & 1.00 & 1.00 & 1.00 & 1.00 & 1.00 & 1.00 \\
S4 & OpenSemiQuant & 0.50 & 0.50 & 1.00 & 0.00 & 0.50 & 1.00 & 1.00 & 1.00 & 1.00 & 0.50 & 1.00 & 1.00 \\
S5 & OpenMaps & 1.00 & 0.00 & 1.00 & 1.00 & 1.00 & 1.00 & 1.00 & 0.00 & 1.00 & 1.00 & 1.00 & 1.00 \\
S6 & ToggleSiMapping & 1.00 & 1.00 & 1.00 & 0.50 & 1.00 & 1.00 & 1.00 & 1.00 & 1.00 & 1.00 & 1.00 & 1.00 \\
S7 & OpenComposite & 1.00 & 0.00 & 1.00 & 1.00 & 0.50 & 1.00 & 1.00 & 1.00 & 1.00 & 1.00 & 1.00 & 1.00 \\
S8 & ReturnToMain & 1.00 & 0.00 & 1.00 & 0.00 & 1.00 & 1.00 & 1.00 & 0.50 & 1.00 & 1.00 & 1.00 & 1.00 \\
\bottomrule
\end{tabular}%
}
\end{table*}

\newpage
\clearpage

\bibliographystyle{unsrtnat}
\bibliography{custom}
\end{document}